\documentclass[lettersize,journal]{IEEEtran}
\usepackage{amsmath,amsfonts}
\usepackage{algorithmic}
\usepackage{algorithm}
\usepackage{array}
\usepackage[caption=false,font=normalsize,labelfont=sf,textfont=sf]{subfig}
\usepackage{textcomp}
\usepackage{stfloats}
\usepackage{url}
\usepackage{verbatim}
\usepackage{graphicx}
\usepackage[dvipsnames]{xcolor}

\def\BibTeX{{\rm B\kern-.05em{\sc i\kern-.025em b}\kern-.08em
    T\kern-.1667em\lower.7ex\hbox{E}\kern-.125emX}}
\usepackage{balance}
\begin{document}
\title{One-Spike SNN: Single-Spike Phase Coding with Base Manipulation for ANN-to-SNN Conversion Loss Minimization}
\author{Sangwoo Hwang, \IEEEmembership{Student Member, IEEE}, Jaeha Kung, \IEEEmembership{Member, IEEE}
\thanks{S. Hwang is with the Department of Electrical Engineering and Computer Science, Daegu Gyeongbuk Institute of Science and Technology (DGIST), Daegu, South Korea.}
\thanks{Jaeha Kung is with the School of Electrical Engineering, Korea University, Seoul, South Korea. J. Kung is the corresponding author (e-mail: jhkung@korea.ac.kr).
}}

\markboth{Under Review}%
{Hwang \MakeLowercase{\textit{et al.}}: One-Spike SNN}
\maketitle

\begin{abstract}
As spiking neural networks (SNNs) are event-driven, energy efficiency is higher than conventional artificial neural networks (ANNs).
Since SNN delivers data through discrete spikes, it is difficult to use gradient methods for training, limiting its accuracy.
To keep the accuracy of SNNs similar to ANN counterparts, pre-trained ANNs are converted to SNNs (ANN-to-SNN conversion).
During the conversion, encoding activations of ANNs to a set of spikes in SNNs is crucial for minimizing the conversion loss.
In this work, we propose a single-spike phase coding as an encoding scheme that minimizes the number of spikes to transfer data between SNN layers.
To minimize the encoding error due to single-spike approximation in phase coding, threshold shift and base manipulation are proposed.
{ Without any additional retraining or architectural constraints on ANNs, the proposed conversion method does not lose inference accuracy (0.58\% on average) verified on three convolutional neural networks (CNNs) with CIFAR and ImageNet datasets.
In addition, graph convolutional networks (GCNs) are converted to SNNs successfully with an average accuracy loss of 0.90\%.
Most importantly, the energy efficiency of our SNN improves by 4.6$\sim$17.3$\times$ compared to the ANN baseline.
}
\end{abstract}


\section{Introduction}
Recent developments in artificial neural networks (ANNs) and their learning algorithms have achieved unprecedented performance improvement in a variety of applications: computer vision, language modeling, and video analytics~\cite{Alexnet,gpt3,video_analytics}.
However, the ANN performance is directly related to the number of parameters, which naturally increases the computation overhead in terms of memory footprint and energy consumption.
On the contrary, running an ANN model on a mobile/edge device requires a better energy efficiency due to limited battery capacity and computing power~\cite{evaluate_energy_ann}.
To improve the energy efficiency, one can come up with a more efficient NN architecture that requires a significantly less number of computations.
An excellent candidate for this is a spiking neural network (SNN)~\cite{nature_roy}.
SNNs are extremely energy efficient since their computations are event-driven as information between layers is transferred by binary spikes with the use of biological neuron models~\cite{izh_model,hh_model}.
To fully exploit the benefit of SNNs, neuromorphic hardware such as Loihi~\cite{loihi}, TrueNorth~\cite{truenorth}, and SpiNNaker~\cite{spinnaker} have been fabricated.

To make the accuracy of SNNs at par with the ANN counterparts, there have been many research efforts in 
training SNNs~\cite{surrogate_train_aaai2021,surrogate_Train_nips2021}.
One of the representative learning methods of SNN is Spike Timing Dependent Plasticity (STDP), which is a biologically plausible learning rule that learns the weights of connections based on the difference in spike timings between neurons.
The STDP rule also can be used to train a Liquid State Machine (LSM) in which neurons are recurrently connected~\cite{Diehl,spilinc,Adaptive-interlink}.   
However, STDP-based learning is an unsupervised learning rule that limits the training accuracy of SNNs on complex tasks.
In order to train SNNs having deep layers in a supervised manner, backpropagation, the most common training method for ANNs, can be utilized for the SNN training as well.
However, due to the nature of non-differentiable spike models in SNNs, effective gradient-based learning could not be applied limiting the maximum achievable accuracy of SNNs~\cite{surrogate_train_aaai2021,surrogate_Train_nips2021}.

Another way to obtain well-performing SNNs is by directly converting pre-trained ANNs.
However, the converted SNNs in prior work suffer from two major limitations: (i) longer timestep~\cite{rmp-snn,TSC}, and (ii) multiple spikes per timestep~\cite{rmp-snn,optimal_deng,spikeconverter,phasecoding,temporal-pattern-coding}.
These limitations lead to high energy consumption making the use of SNNs less promising.
To achieve high accuracy with fewer timesteps, some prior works fine-tune SNNs after the conversion with additional training iterations~\cite{diet-snn,onetimestep_allyouneed}.
Other previous studies have modified the structure of the ANN, e.g., removing batch normalization layers, in order to reduce the conversion loss before training the ANN (i.e., conversion-aware training~\cite{pre_training_iclr,pretraining_ijcai}).
These approaches, however, are only applicable to a limited number of ANN models due to the structural constraints.


In this paper, we propose a highly efficient ANN-to-SNN conversion method that minimizes the number of spikes in the converted SNN, named {\it one-spike SNN}, with negligible accuracy loss. 
The contributions of this paper are as follows:
\begin{enumerate}
    \item \textbf{Generality of conversion process}: Our work converts any ReLU-based ANNs to SNNs without constraints on batch normalization or pooling layers. Moreover, no post-training iterations are required after the conversion. The generality of conversion is verified by converting multiple CNN and GCN models on various datasets.
    \item \textbf{High energy-efficiency}: By allowing only one spike per neuron, the power consumption of updating the membrane potential of a neuron becomes minimal. 
{    
The power reduction compared to ANNs is limited in SNNs using multiple spikes per neuron. 
Multiple spikes translate to more computations required to update the membrane potential of neurons.
In contrast, our conversion is highly power-efficient as it only allows a single spike per neuron (Section~\ref{sec:single-spike}). 
Compared to the prior works that use one spike per neuron, i.e., temporal coding, the required timestep is minimized by applying the proposed base manipulation technique (Section~\ref{sec:reduce_loss}), resulting in a significant reduction in energy consumption.
}
    \item \textbf{Low conversion loss}: We present two novel error reduction methods, i.e., threshold shift and base manipulation, to minimize the conversion loss when the proposed single-spike approximation is used.
{   
We theoretically set the optimal base value that minimizes the conversion error for a given timestep.    
    
}
\end{enumerate}

\section{Preliminaries}
\subsection{Activation Encoding in ANN-to-SNN Conversion}

\begin{figure}[]
    \centering 
    \includegraphics[scale=0.375]{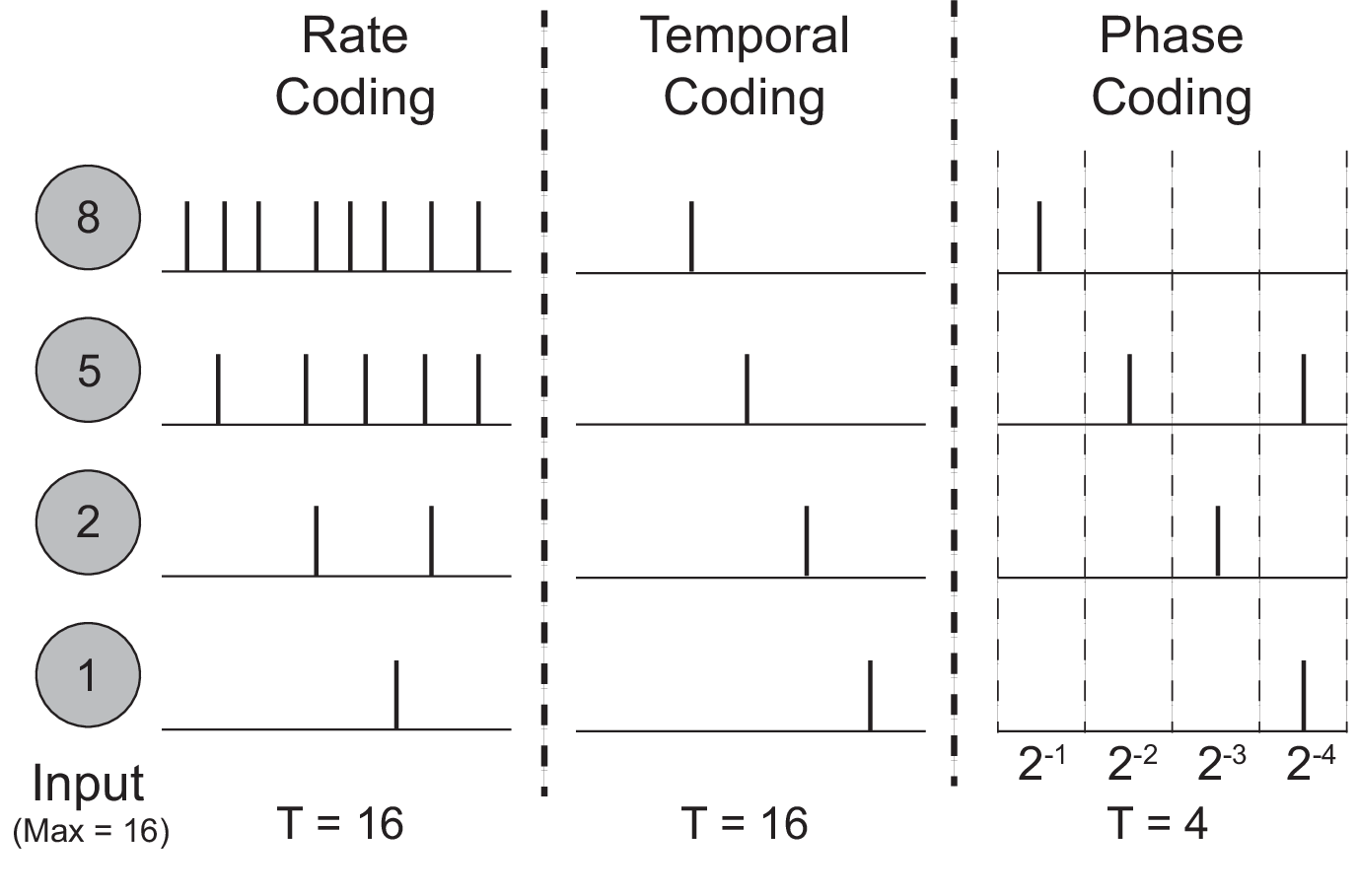}
    \caption{Various activation encoding schemes: rate coding, temporal coding, and phase (or binary) coding. In this example, base 2 is used for illustrating the phase coding.}
    \label{fig:encoding}\vspace{-2mm}
\end{figure}

When converting an ANN to an SNN, the conversion accuracy highly depends on the encoding of activation values to a set of spikes (Fig.~\ref{fig:encoding}).
Most frequently used encoding scheme is rate coding, which generates spikes at each neuron by Poisson sampling during a pre-defined discrete timestep $T$~\cite{rmp-snn,optimal_deng,spikeconverter,layer-norm}.
The ratio of the current activation value and the maximum value becomes the sampling probability.
For complex vision tasks, e.g., ImageNet classification, longer timestep $T$ (higher resolution) is required to achieve high classification accuracy.
Unfortunately, a large activation value converts to a higher number of spikes, which increases the processing time and power consumption of SNN hardware.

Unlike the rate coding, temporal coding produces only one spike at each timestep $T$.
With time-to-first-spike (TTFS) coding~\cite{TSC,TTFS}, spike time $t_i$ at which spike is produced within timestep $T$ implies the activation value $x_i$ at the $i$\textsuperscript{th} neuron. 
Using an exponential decaying TTFS scheme~\cite{t2fsnn}, `$e^{\frac{T-t_i}{T}}$' is proportional to `$x_i/x_{max}$', where $x_{max}$ is the maximum activation value ($x_{max}=16$ in Fig.~\ref{fig:encoding}).
Thus, a neuron having a large activation value produces a spike at earlier time than a neuron with a small activation value.
As the temporal coding allows only one spike at each timestep $T$, spike rate is much lower than an SNN with the rate coding resulting in higher energy efficiency. 
However, to ensure little accuracy loss in approximating the activation value, temporal coding generally requires a longer timestep than the rate coding.

The other coding scheme is phase (or binary) coding that combines the advantages of both rate and temporal coding~\cite{phasecoding,temporal-pattern-coding,logarithimic_coding}.
The phase coding assigns different weights $w_t$ to each time segment $t$ within $T$ so that activation values are encoded by both the number of spikes and the spike pattern.
In binary coding, i.e., a special case of the phase coding, $w_t$ is determined by powers of 2 (base = 2).
Since multiple spikes are allowed, e.g., $x_i/x_{max}$ is proportional to $\sum_t 2^t/2^T$, a shorter timestep $T$ is required than the temporal coding.
Thanks to the use of varying weights $w_t$, fewer spikes are needed to represent an activation value than the rate coding.
In this paper, we maximize the efficiency of phase coding by allowing only one spike within $T$ while approximation error is minimized by adjusting the base $Q$ smaller than 2.


\subsection{Neuron Model}\label{sec:neuron_model}
A neuron model in an SNN takes over the role of an ANN's (nonlinear) activation function. 
Similar to other ANN-to-SNN conversion works, we select a leaky integrate-and-fire (LIF) model to emulate a well-known rectified linear unit (ReLU).   
{
According to \cite{izh_model}, a LIF model has the simplest biological plausibility but requires only 5 floating-point operations, while Hodgkin-Huxley~\cite{hh_model} has the best biological plausibility but with 1,200 floating-point operations. 
Since the goal of converting an ANN to an SNN is to maximize energy efficiency, the LIF model becomes the natural choice.}
The LIF neuron is defined as:
\begin{equation} 
\begin{split}
    v_{j}(t) =  \lambda_jv_{j}(t-1) +\sum_{i=1}^{N_{pre}}w_{ij}s_{i}(t-1)-V_{\theta}^js_{j}(t),
\\ 
    s_{j}(t)=\begin{cases}
			1, & \text{$v_j(t)>V^{j}_{\theta}$}\\
            0, & \text{otherwise}
		 \end{cases},
\end{split}\label{eq:lif}
\end{equation}
where $i$ or $j$ is the index of a pre- or post-synaptic neuron, $N_{pre}$ is the number of pre-synaptic neurons,
$v_j$ is the membrane potential of the post-synaptic neuron $j$, 
and $\lambda$ is the leak factor of the membrane potential. 
The $w_{ij}$ is the strength of a synaptic connection
between the neuron $i$ and $j$, $s_{i}(t)$ is the spike ($0$ or $1$) of the pre-synaptic neuron $i$ at time $t$.
The neuron's membrane potential $v_j(t-1)$ changes by factor $\lambda_j$ and weighted sum of pre-synaptic spikes increases the potential. 
If $v_j(t)$ becomes higher than the threshold $V^{j}_{\theta}$, a neuron $j$ fires a spike to its post-synaptic neurons and $v_j(t)$ may reset to zero (hard reset). 
Instead of the hard reset, we reduce $v_j(t)$ by the firing threshold (soft reset), since it mimics the ReLU function better~\cite{rmp-snn}.

\begin{figure}
    \centering 
    \includegraphics[scale=0.34]{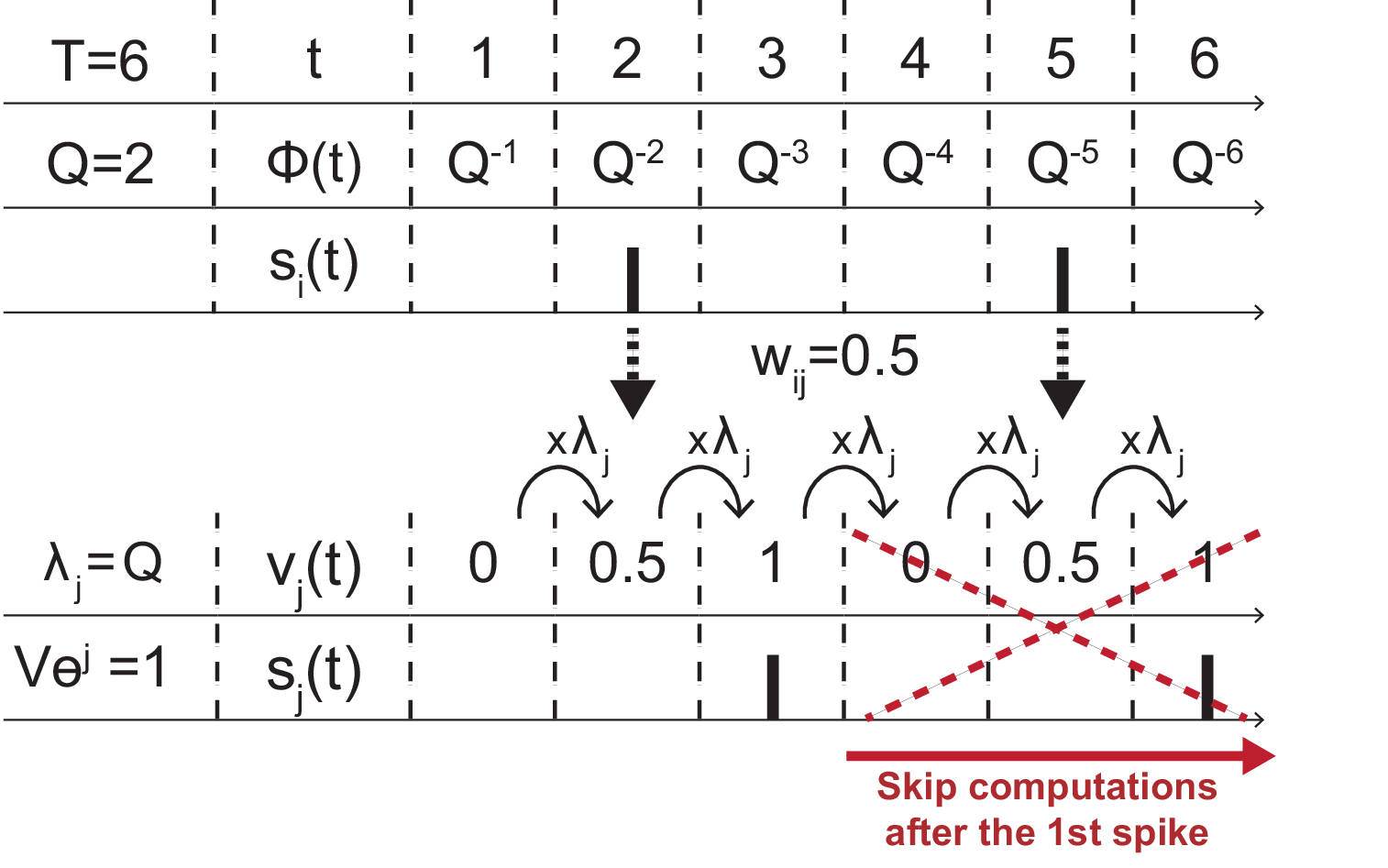}
    \caption{Operation of a neuron model in the proposed one-spike SNN. A single spike phase coding is used at hidden layers with base ($Q$) manipulation. Each spike at $t$ has a phase $Q^{-t}$ where $Q=2$ for binary coding.}
    \label{fig:phase_coding_example}
\end{figure}

There are two ways to implement the phase coding: (i) spikes are multiplied by the weight $w_{ij}(t)$ corresponding to the current phase $\phi(t)$ (synapse-based)~\cite{phasecoding}, and (ii) $\lambda_j$ is multiplied to the membrane potential (neuron-based)~\cite{temporal-pattern-coding,logarithimic_coding}.
The latter is used to realize the phase coding in this paper.
The difference between the prior work and our model is that we propose to use base $Q$ that is other than 2 so that ANN-to-SNN conversion loss is minimized (Section~\ref{sec:reduce_loss}).
Fig.~\ref{fig:phase_coding_example} illustrates the operation of the neuron model based on Eq.~(\ref{eq:lif}) when binary coding is used ($Q=2$).
The membrane potential $v_j(t)$ is multiplied by $\lambda_j=Q$ when the phase $t$ changes from 1 to 2.
This has the same effect of multiplying $Q^{-2}$ to the spike $s_i(t)$ fired at $t=2$.
Whenever $s_j(t)$ reaches the $V_\theta^j$, the post-synaptic neuron fires a spike (at $t=3$ in Fig.~\ref{fig:phase_coding_example}).
As will be discussed in Section~\ref{sec:single-spike}, we skip computations at later phases when the first spike occurs ({\it single-spike approximation}).


\subsection{Weight Normalization}\label{sec:weight_normal}
To minimize the ANN-to-SNN conversion loss, each neuron requires a proper threshold. 
If the threshold is much higher than the weighted sum, neurons never fire and the total firing rate converges to zero as layer gets deeper.
On the other hand, if the threshold is much lower than the weighted sum, neurons always fire and no useful information can be extracted.
Keeping the ratio between the weights and the threshold within a reasonable range is called `weight normalization'. 

\begin{figure}
    \centering 
    \includegraphics[scale=0.31]{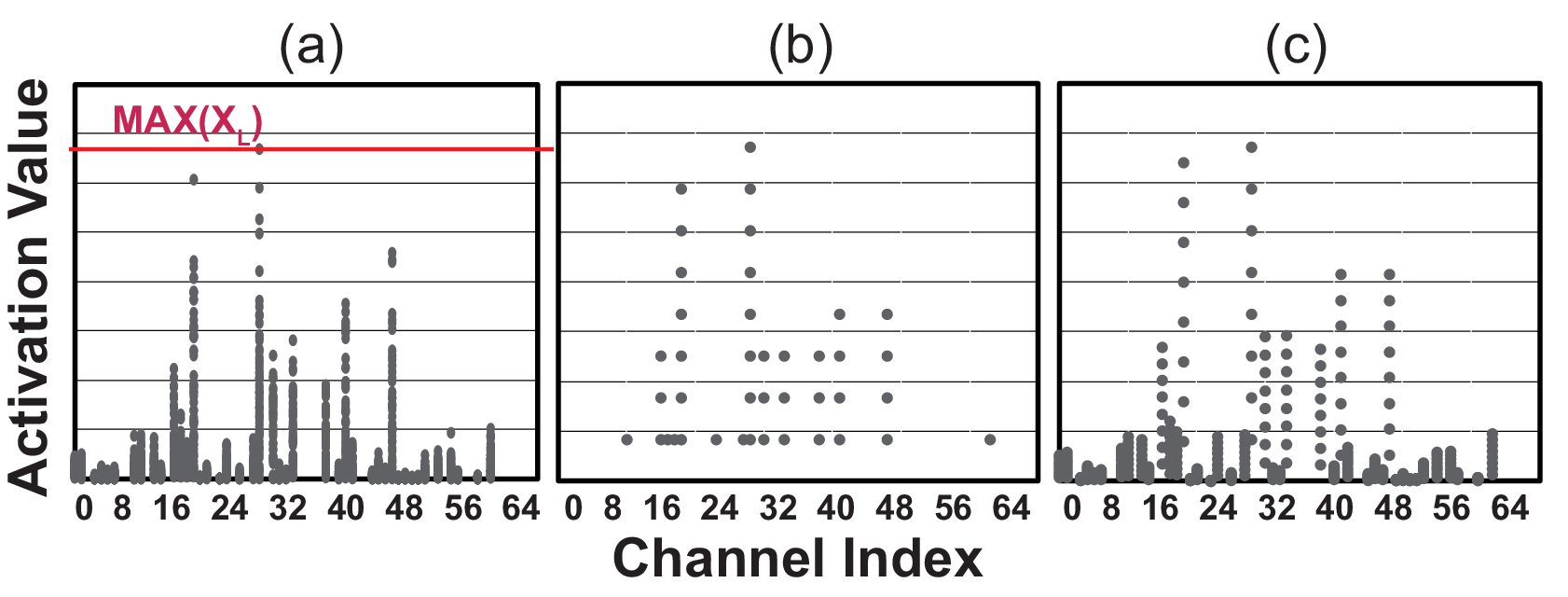}
    \caption{{Activation values at each channel at the first convolution layer of VGG-16 for ImageNet classification. (a) Original activation values of the ANN (i.e., VGG-16). Encoded values of spikes in the converted SNN (b) with layer normalization, and (c) with channel-wise normalization.}}
    \label{fig:normalization}
\end{figure}

The most common weight normalization method is layer normalization~\cite{layer-norm}, which normalizes weights $w^l_{ij}$ at layer $l$ by each layer's maximum activation value `${\max}(\textbf{X}_{l})$'.
For an ANN with many channels, channel-wise normalization performs better than the layer normalization~\cite{channel-wise_norm}.
{
To verify the difference between the two normalization methods, we provide the activation distribution in the first layer of VGG-16 trained on ImageNet in Fig.~\ref{fig:normalization}(a).
Even in the same layer, the maximum activation value at each channel significantly varies. 
Channels with significantly smaller activations than the layer's maximum activation value can be poorly represented by spikes when layer normalization is used. 
In the case of the rate coding, for instance, each spike represents a value of ${\max}(\textbf{X}_{l})/T$, as shown in Fig.~\ref{fig:normalization}(b). 
Due to the large step size, i.e., ${\max}(\textbf{X}_{l})/T$, the layer normalization incurs distribution mismatch after the conversion.
On the other hand, suppose that spikes in each channel are normalized to their own ${\max}(\textbf{X}^{c}_{l})$ where $c$ indicates the channel index. 
In this case, the step size becomes small for channels that have a narrow distribution.
As a result, the channel-wise normalization significantly reduces conversion errors matching the original distribution, as shown in Fig. \ref{fig:normalization}(c).
}
The channel-wise normalization is defined as:
\begin{equation} 
w_{ij} = \frac{{w}_{ij}\cdot{\max}(\textbf{X}_{l-1}^{C_i})}{{\max}(\textbf{X}_{l}^{C_j})},  \hspace{2mm}
{b}_{j} = \frac{{b}_{j}}{{\max}(\textbf{X}_{l}^{C_j})},
\end{equation}
where $i$ or $j$ is the index of a pre-/post-synaptic neuron, $\textbf{X}$ is the input activation, $l$ is the current layer index, $C_i$ or $C_j$ is the input/output channel index, and $b_{j}$ is the bias.
As the pre-synaptic spikes are multiplied by the normalized weight at layer `$l-1$', we need to void this normalization and perform a new weight normalization by `${\max}(\textbf{X}_{l}^{C_j})$'.
On the other hand, the bias term is not affected by weights of the previous layer, thus it is simply normalized by the maximum activation value.
We use the layer-wise normalization for fully-connected layers or pooling layers where channels do not exist.


\section{Activation Encoding in One-spike SNN}\label{sec:proposed_enc}

\subsection{Single-Spike Phase Coding}\label{sec:single-spike}

The main advantage of SNNs is that they require only additions whenever a pre-synaptic spike is observed, i.e., $\sum w_{ij}s_{i}(t-1)$ in Eq.~(\ref{eq:lif}), instead of multiply-accumulate (MAC) operations in ANNs.
The energy consumption of a 32-bit MAC unit is 5$\times$ higher than that of a 32-bit adder~\cite{horowitz}.
However, if the number of spikes generated is significantly large within a timestep $T$, the energy consumption becomes similar to the ANN model.
Like the temporal coding, if only one spike occurs per timestep, the energy efficiency can be maximized.
Therefore, the conventional phase coding, which produces multiple spikes, is less energy efficient than the temporal coding.

Instead of allowing multiple spikes in the binary/phase coding, we propose to limit the number of spikes to one for activations in hidden layers ({\it single-spike approximation}; Fig.~\ref{fig:phase_coding_example}) so that energy efficiency becomes at par with the temporal coding.
When converting the input data, e.g., image pixels, multiple spikes are allowed as in the original phase coding to maintain the accuracy.
As presented in Fig.~\ref{fig:phase_coding_example}, only one generated spike $s_j(t)$ at the earliest phase ($t=3$) fires which has the largest value that approximates the true activation value.
Since there is only one spike that fires from neuron $j$, the number of additions that are required at the following neurons significantly reduces.
However, this comes at the cost of a larger encoding error than the original multi-spike phase coding.
For instance, 12 ($1010$) and 11 ($1001$) are encoded as 8 ($1000$), and 7 ($0111$) and 6 ($0110$) are encoded as 4 ($0100$) when $Q=2$.
These encoding errors propagate through deep layers resulting in incorrect decisions at the output of the converted SNN model.
To minimize the error due to single-spike approximation, we propose two schemes: (i) threshold $V_\theta$ shift, and (ii) manipulation of base $Q$.

\subsection{Minimizing Conversion Error}\label{sec:reduce_loss}
\subsubsection{Threshold Shift for Round-off Approximation}\label{sec:round_off}
\begin{figure*}[!t]
    \centering 
    \includegraphics[scale=0.42]{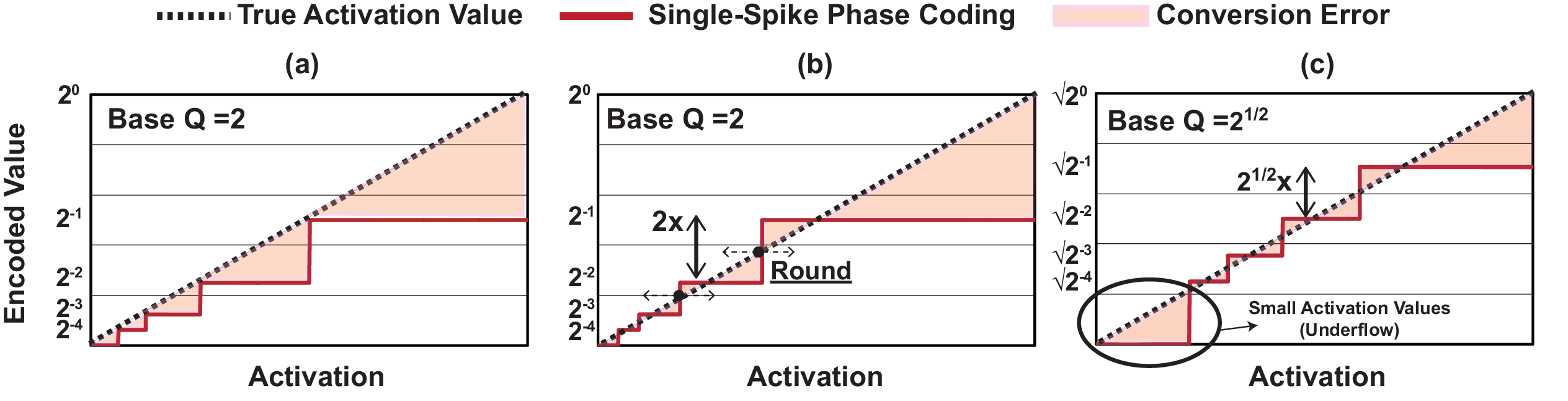}
    \caption{Conversion errors when different single-spike phase coding schemes with timestep $T=4$ are used: (a) nai\"ve single-spike binary coding ($Q=2$), (b) single-spike binary coding with round-off approximation, and (c) single-spike phase coding ($Q<2$; base manipulation) with round-off approximation.}
    \label{fig:conversion_loss}\vspace{-2mm}
\end{figure*}

By allowing only one spike in phase coding with timestep $T$, the number of values that can be expressed is limited to $T$.
Each representable value equals to $Q^{-t}$ at phase $t$. 
Since the following spikes are ignored with the single-spike approximation, all activation values greater than $Q^{-t-1}$ and less than $Q^{-t}$ are floored to $Q^{-t-1}$, i.e., conversion error.
The estimated encoded value with the binary coding using a na\"ive single-spike approximation is defined as:
\begin{equation}\label{eq:naive-single-spike} 
    \Bar{x}=\begin{cases}
    0,& x< Q^{-T}\\
    Q^{\lfloor log_{Q}{x} \rfloor},&  Q^{-T} \leq x < Q^{-1}\\
    Q^{-1}, &  Q^{-1}\leq x
    \end{cases},
\end{equation}
where $\Bar{x}$ is the encoded value of the true activation value $x$ $\in$ [0,1), $T$ is the timestep, and $Q$ is $2$.
Fig.~\ref{fig:conversion_loss}(a) shows an example of how the conversion error looks like using the na\"ive single-spike approximation when $T=4$.
An activation value closer to $Q^{-t}$ has a larger error due to the floor operation.
Thus, the maximum error that can happen at phase $t$ becomes `$(Q^{-t}-Q^{-t-1})$'.

If we approximate the activation value by rounding off rather than the floor operation, the worst-case approximation error reduces by half, i.e., `$(Q^{-t}-Q^{-t-1})/2$' (Fig.~\ref{fig:conversion_loss}(b)).
For the round-off operation, a threshold should be set to a midpoint between two consecutive phase values, i.e., `$({Q^{-t} + Q^{-t-1}})/{2}$'.
If the activations are larger (or smaller) than the threshold, they round to $Q^{-t}$ (or $Q^{-t-1}$).
The encoded value $\Bar{x}$ with the single-spike approximation using rounding operation can be represented as:
\begin{equation}\label{eq:round-off} 
    \Bar{x}=\begin{cases}
    0,& x< Q^{-T}\frac{Q+1}{2Q}\\
    Q^{\lfloor log_{Q}{\frac{2Q}{Q+1}x} \rfloor},&  Q^{-T}\frac{Q+1}{2Q} \leq x < Q^{-1}\\
    Q^{-1}, &  Q^{-1}\leq x
    \end{cases}.
\end{equation}
In this case, the closer the value is to the threshold, the larger the approximation error.
The round-off approximation is realized by {\it shifting the firing threshold} $V_\theta$ to $\frac{1+1/Q}{2}$ instead of 1 in the example shown in Fig.~\ref{fig:phase_coding_example}.
Note that the threshold determines when a neuron to fire a spike.
As the membrane potential of the neuron $j$, i.e., $v_j(t)$, is multiplied by $\lambda_j=Q$ at the next phase, this has the same effect of dividing the $V_\theta$ by $Q$.
Therefore, at $t=2$, the threshold $V_\theta$ effectively becomes $({Q^{-1} + Q^{-2}})/{2}$.

\subsubsection{Manipulation of Base Q}\label{sec:manipulation_base}

\begin{figure}
    \centering 
    \includegraphics[scale=0.36]{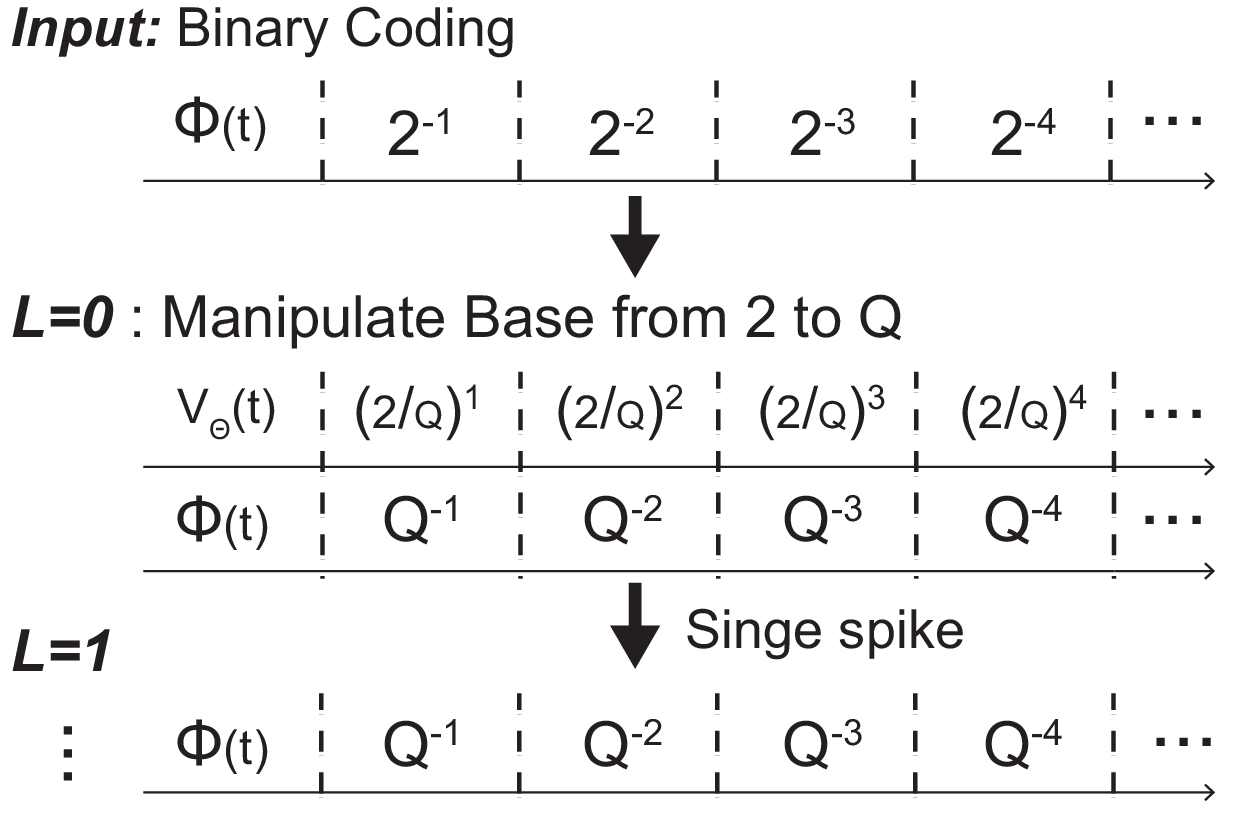}
    \caption{An example of manipulating base $Q$ in our single-spike phase coding. Here, external input is binary coded ($Q=2$) while the remaining layers ($L=0,1,...$) have the base $Q$ less than 2. This is done by scaling the threshold $V_\theta$ of layer $L=0$ by ${2}/{Q}$.}
    \label{fig:mainpulation_base}
\end{figure}

Fig.~\ref{fig:conversion_loss}(a-b) show the conversion errors when binary coding is used ($Q=2$).
For both cases, the activation encoding error due to the single-spike approximation is proportional to `$({Q^{-t}-Q^{-t-1}})$'.
Thus, it becomes natural to use a smaller base $Q$ to reduce the conversion error.
As shown in Fig.~\ref{fig:conversion_loss}(c), the smaller the Q value ($Q=\sqrt{2}$), the smaller the conversion error for large activation values, which has higher contribution to the ANN output.
As a result, the approximation errors reduce by $\sqrt{2}\times$ when $Q=\sqrt{2}$ compared to the binary coding. 
When converting the (external) input data, we do not change $Q$ and multiple spikes are allowed (i.e., binary coding).
To manipulate the $Q$ value to become less than 2 in the following layers, we change the neuron threshold of the first hidden layer ($L=0$ in Fig.~\ref{fig:mainpulation_base}).
In order for the output spikes to have a different Q value, we scale the threshold by $\frac{2}{Q}$ at each phase $t$.
With this modification, output spikes are generated with respect to the base $Q$.
In Fig.~\ref{fig:mainpulation_base}, the base has changed from 2 to $Q$ at $L=0$, but this can be generalized to any layers and any base values.
When the base of incoming spikes of the $L$\textsuperscript{th} layer is $Q_{L-1}$, the base of output spikes can be manipulated to $Q_L$ by scaling its threshold by $\frac{Q_{L-1}}{Q_L}$.

However, reducing $Q$ does not minimize the total conversion error since a small base $Q$ limits the representable range of activations.
The representable value range is $[Q^{-T},Q^{-1}]$ with base $Q$ using timestep $T$.
As $Q$ decreases to near 1, the range of activation values that can be approximated becomes extremely narrow.
As presented in Fig.~\ref{fig:conversion_loss}(c), small activation values that are less than $Q^{-T}$ becomes zero (underflow).
In other words, neurons with small activation values seldom fire in the converted SNN with a small base $Q$.
Moreover, activation values follow a normal distribution implying that a large amount of activations may become zero leading to significant ANN-to-SNN conversion loss.
Allowing a longer timestep $T$ helps reduce the underflow error at the cost of latency.
Thus, setting appropriate base $Q$ and timestep $T$ is crucial in minimizing the conversion loss when using the proposed single-spike phase coding.

\begin{figure}[t]
    \centering 
    \includegraphics[scale=0.4]{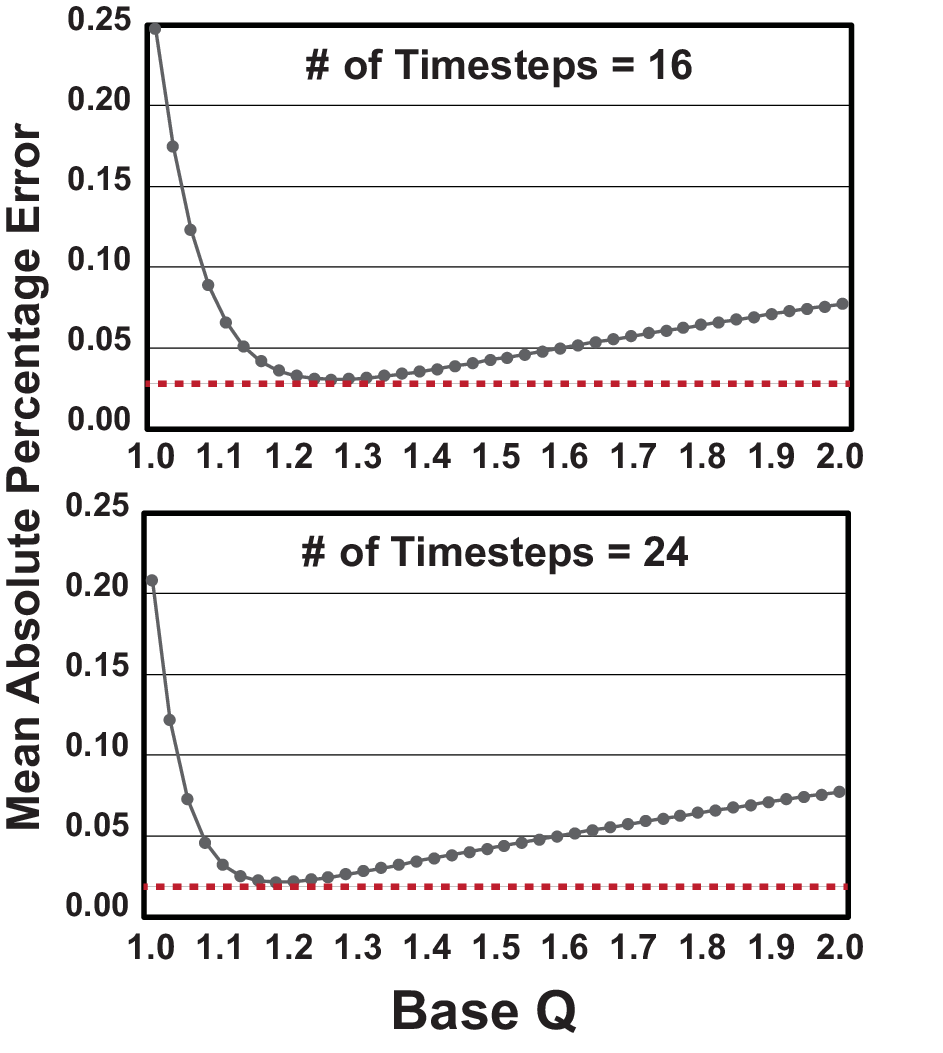}
    \caption{{The computed mean absolute percentage error in converting activation values (normal distribution) with the single-spike approximation at various base $Q$ values (1.0$\sim$2.0) and the number of timesteps $T$ (16 and 24).}}
    \label{fig:optimal_q}
\end{figure}

{

Assuming that the activation values are Gaussian distributed as a result of batch normalization in CNNs, we can derive the conversion loss from the proposed single-spike approximation at a fixed timestep $T$ and a given base $Q$.
We analyzed the percentage error of the encoded spike to the original activation value since even a small activation can significantly affect the final output as it may propagate to the following layers through strong weights.
Therefore, the conversion error is estimated by the mean absolute percentage error~(MAPE) and can be expressed as:
\begin{equation}\label{eq:loss_function}
\begin{split}
        \int^{1}_{0}|\frac{x-\Bar{x}}{x}|f(x)dx\\
    =\int^{1}_{0}|\frac{x-\Bar{x}}{x}|\frac{1}{\sigma\sqrt{2\pi}}e^{-\frac{1}{2}{(\frac{x-\mu}{\sigma}})^2}dx, 
\end{split}
\end{equation}
where $\Bar{x}$ is the encoded value according to Eq.~(\ref{eq:round-off}) and $f(x)$ is the probability density function of Gaussian distributed activations.
The $\mu$ and $\sigma$ are the mean and the standard deviation of activation values, respectively. 
With $\mu=0$ and $\sigma=1$, Fig.~\ref{fig:optimal_q} shows the computed MAPE at various base $Q$ values (1.0$\sim$2.0) and timesteps (16 and 24).
When $T=16$, the minimum MAPE is achieved when $Q$ is near 1.3; when $T=24$, $Q=1.2$ results in the minimum MAPE. 
This analysis, however, is not exact since the batch normalization of each layer provides different $\mu$ and $\sigma$ values.
Still, it provides a good starting point to empirically search the optimal base $Q$.
In Section~\ref{sec:impact_q}, we show that the optimal base $Q$ computed by Eq.~(\ref{eq:loss_function}) and the empirically found optimal $Q$ are closely matched.  
}

\subsection{Neuron Model in One-spike SNN for CNN Conversion}\label{sec:CNN-to-SNN}
The SNN model obtained by ANN-to-SNN conversion with the proposed activation encoding is named {\it one-spike SNN}.
The proposed single-spike approximation and novel error reduction methods are described in Algorithm~\ref{alg:neuron_model} for convolution layers.
In this work, we can implement batch normalization and bias of each channel $C_j$ in SNNs by properly initializing the potential $v_j(t)$ and the threshold $V_\theta^j$.
Since the generated spike is the result of dividing the membrane potential by the threshold, the batch normalization of each channel  
can be fused into the threshold (line 4).
Also, the bias can be considered in the membrane potential prior to the inference (line 5).
Another approach is to directly fuse the batch normalization parameters into weights, but the distribution of the weights becomes wider, which increases the number of bits required to quantize the weights.
{ 
In addition, the rounding approximation (Section~\ref{sec:round_off}) shifts the threshold to the midpoint `$\frac{Q+1}{2Q}$' between two consecutive phases.
According to the neuron model in the phase coding, the potential $v_j(t-1)$ is multiplied by $Q_{L-1}$ and the weighted sum of the pre-synaptic spikes, $\sum_{i=1}^{N_{pre}} w_{ ij}\cdot s_{i}(t-1)$ is added to the potential over timestep $T$ (i.e., iterate $T$ times).
At each timestep, adjusting the threshold by $\frac{Q_{L-1}}{Q_{L}}$ makes the base of the output spike to $Q_{L}$ (Section~\ref{sec:manipulation_base}).
Finally, if the $v_j(t)$ exceeds the threshold, a neuron $j$ propagates the spike to its post-synaptic neurons, and the computation ends.

The proposed neuron model is also used for other non-convolution layers (i.e., max pooling, avg pooling, and linear layers).
}
Similar to the temporal coding, we implement max pooling layers by skipping the computations of other neurons when a neuron in the same group generates the first spike.
At the output layer, there is no need to generate spikes since no layers are followed by the output layer.
After updating the membrane potential of output neurons, classification is performed by identifying the neuron index with the maximum potential.


\begin{algorithm}[t]
\caption{Neuron model for convolution layer}
\begin{algorithmic}[1]\label{alg:neuron_model}

\STATE{Number of total timestep: $T$} 
\STATE{Channel index of neuron $j$ : $C_j$}
\STATE{ (0) Initialization of threshold and potential for neuron $j$}
\STATE{$V_{\theta}^{j} \gets \frac{\sqrt{{\sigma_{C_j}}^{2}}}{\gamma_{C_j}}\frac{Q+1}{2Q}$},  
\\
\STATE$v_j(t) \gets \frac{\beta_{C_j}\frac{\sqrt{{\sigma_{C_j}}^{2}}}{\gamma_{C_j}}-\mu_{C_j}}{{\max}(\textbf{X}_{l}^{C_j})}$
\WHILE{$t < T$}
    \STATE{ (1) Update membrane potential of neuron $j$}
    \STATE{$v_j(t)\gets v_j(t-1)Q_{L-1}$}
    \STATE{$v_j(t) \gets v_j(t)+\sum_{i=1}^{N_{pre}}w_{ij}s_{i}(t-1)$} 
    \STATE
    \STATE{ (2) Manipulate base $Q$ by modifying the threshold}
    \STATE{$ V_{\theta}^{j} \gets V_{\theta}^{j}\frac{Q_{L-1}}{Q_{L}}$}
    \IF{$v_j(t)>V_{\theta}^{j}$}
        \STATE{ $s_{j}(t) \gets 1$}
        \STATE{\textbf{break}}
    \ELSE
        \STATE{ $s_{j}(t) \gets 0$}
    \ENDIF


\ENDWHILE
\end{algorithmic}
\end{algorithm}

\begin{figure*}
    \centering 
    \includegraphics[scale=0.275]{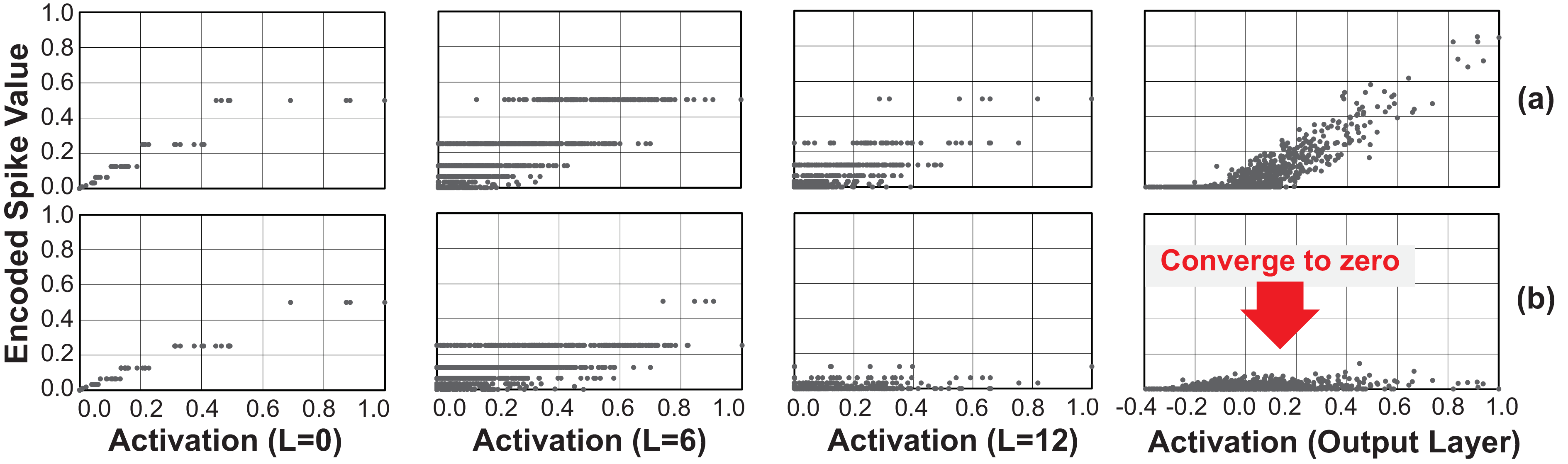}
    \caption{Scatter plots showing the relationship between the true activation values in ANN (x-axis) and the encoded spike values in converted SNN (y-axis) with the single-spike approximation at VGG16 for ImageNet Classification: (a) approximation with $V_\theta$ shift, and (b) approximation without $V_\theta$ shift.}
    \label{fig:round_off}\vspace{-2mm}
\end{figure*}


\section{Experimental Results}\label{sec:result}
\subsection{Experimental Setup}

The proposed one-spike SNN is implemented in CUDA within the PyTorch framework.
As benchmarks, pre-trained VGG-16, ResNet-18 and ResNet-34 models on CIFAR-10, CIFAR-100 and ImageNet are directly converted to one-spike SNNs.
The major difference between the one-spike SNN and the prior work \cite{pre_training_iclr,pretraining_ijcai,t2fsnn_hardware} is that the conversion process does not require any constraints on ANNs or conversion-aware ANN training.
In~\cite{rmp-snn,diet-snn}, conversion was performed by removing batch normalization layers or bias terms resulting in lower accuracy.
For demonstrating the effectiveness of direct ANN-to-SNN conversion, we have selected public pre-trained models.

\subsection{Accuracy of One-Spike SNN}\label{sec:accuracy}
\subsubsection{Impact of Threshold Shift}

The impact of the threshold shift for round-off approximation is demonstrated by comparing conversion error at several intermediate layers and the output layer.
Fig.~\ref{fig:round_off} presents encoded activations with the single-spike approximation and their true activation values in the ANN.
The Fig.~\ref{fig:round_off}(a) shows the encoding result with the threshold shift and Fig.~\ref{fig:round_off}(b) without the threshold shift.
There are small flooring errors at early layers, e.g., $L=0$, when encoding is performed without the threshold shift.
However, as the layer gets deeper, these flooring errors accumulate and output values mostly converge to zero.
As can be seen at $L=12$ in Fig.~\ref{fig:round_off}(b), most of neurons do not fire implying that no information is delivered.
On the other hand, the activation encoding with round-off (Fig.~\ref{fig:round_off}(a)) shows the linear relationship with the true output values at the output layer (negligible accuracy loss). 

\subsubsection{Impact of Base Q}\label{sec:impact_q}
\begin{figure}
    \centering 
    \includegraphics[scale=0.35]{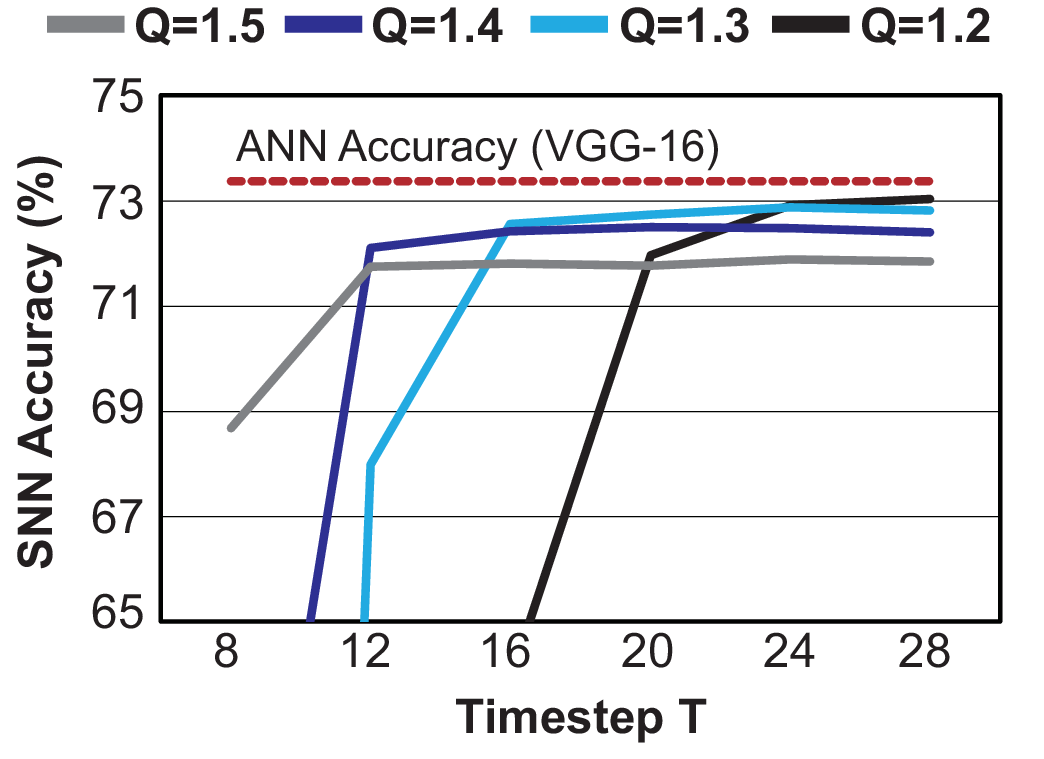}
    \caption{Accuracy of one-spike SNN on ImageNet classification by varying the base $Q$ and timestep $T$.}
    \label{fig:accuracy_by_Q}
\end{figure}
As explained in Section~\ref{sec:reduce_loss}, as the base $Q$ gets closer to 1, the single-spike approximation becomes more precise within the representable value range.
However, having a smaller $Q$ requires a longer timestep due to the increased underflow errors for activations smaller than $Q^{-T}$ (Fig.~\ref{fig:accuracy_by_Q}).
For instance, when $Q=1.2$, the timestep should be higher than 20 to maintain the SNN accuracy (e.g., $T=24$).
As more low-valued activations become zero with a smaller $Q$, the accumulated underflow error significantly drops the SNN accuracy.
Thus, increasing $T$ broadens the value range since the minimum representable value equals to $Q^{-T}$.
In other words, there is a trade-off between the maximum achievable accuracy (small $Q$) and the inference speed (short $T$).
\begin{figure}
    \centering 
    \includegraphics[scale=0.5]{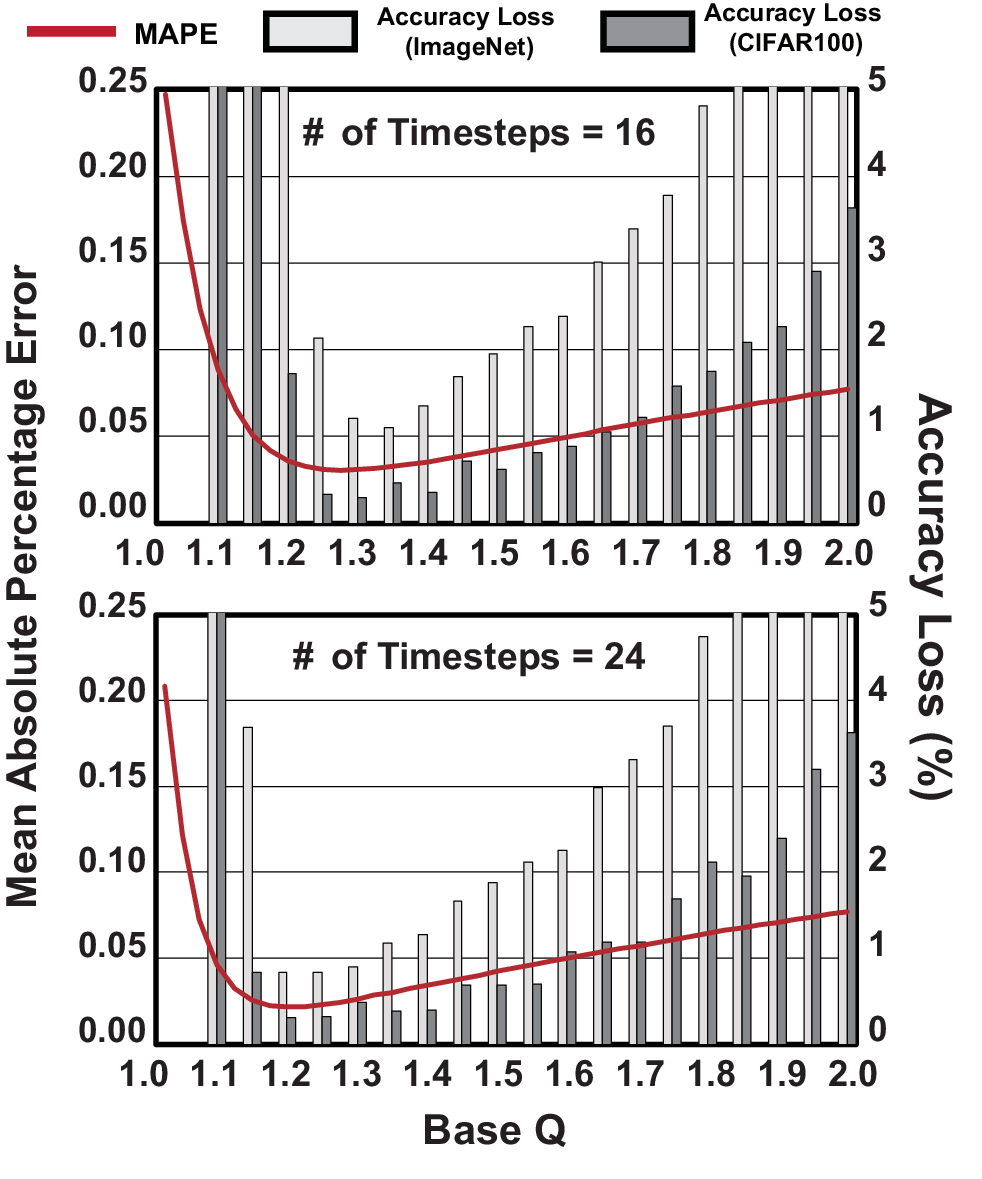}
    \caption{{The mean absolute percentage error of the proposed one-spike SNN and the accuracy loss on image classification tasks (CIFAR-100 and ImageNet) at various base $Q$ values at a given timestep $T$.}}
    \label{fig:emprical}
\end{figure}

{
To achieve the minimum conversion loss with short $T$, we utilized Eq.~(\ref{eq:loss_function}) to select the optimal $Q$ at a given timestep $T$, as shown in Fig.~\ref{fig:optimal_q}. 
Fig.~\ref{fig:emprical} presents the MAPE and accuracy loss according to the base $Q$ values when converting a pre-trained VGG-16 to an SNN for CIFAR-100 and ImageNet.
It validates that the MAPE works well as a metric to measure the quality of $Q$ on the final accuracy.
The accuracy loss follows the same trend as the MAPE; if $Q$ is too small, the accuracy loss significantly increases as underflow occurs. 
Therefore, if the timestep $T$ is fixed, the optimal $Q$ can be selected by Eq.~(\ref{eq:loss_function}) regardless of the network's structure or the dataset.
For the following experiments, we select $Q=1.3$ for $T=16$ and $Q=1.2$ for $T=24$.
}

\subsubsection{Performance on Image Classification}

Table~\ref{tab:cifar10}-\ref{tab:Imagenet} compare the classification accuracy and the required timestep between the prior work and the proposed one-spike SNN (denoted as `\textbf{Ours}' in the table). 
Our proposed method enables accurate ANN-to-SNN conversion with a shorter timestep $T$ with only one spike per neuron (energy-efficient).
The selected base $Q$ is reported in parentheses as a reference.

\begin{table}[!t]
\caption{Comparisons on Accuracy and Required Timestep on CIFAR-10}
\label{tab:cifar10}
\resizebox{\columnwidth}{!}{%
\begin{tabular}{|lllll|}
\hline
\multicolumn{1}{|l|}{\textbf{Model}} & \multicolumn{1}{l|}{\textbf{Coding}} & \multicolumn{1}{l|}{\textbf{ANN (\%)}} & \multicolumn{1}{l|}{\textbf{SNN (\%)}} & \multicolumn{1}{l|}{\textbf{$T$}}  \\ \hline
\multicolumn{5}{|c|}{\textbf{VGG-16}} \\ \hline
\cite{rmp-snn} & Rate & 93.63 & 93.63 & 2048  \\
\cite{TSC} & Temporal & 93.63 & 93.63 & 2048  \\
\cite{t2fsnn} & Temporal & - & 91.43 & 80  \\
\cite{optimal_deng} & Rate & 92.34 & 92.24 & 128  \\
\cite{pre_training_iclr}$^\dag$ & Rate & 95.40 & 95.54 & 32  \\
\cite{spikeconverter} & Rate & 93.71 & 93.63 & 16  \\
\textbf{Ours ($Q=1.3$)} & \textbf{Phase} & \textbf{93.90} & \textbf{93.76} & \textbf{16}  \\ \hline
\multicolumn{5}{|c|}{\textbf{ResNet-18}} \\ \hline
\cite{pre_training_iclr}$^\dag$ & Rate & 96.04 & 96.08 & 32  \\
\textbf{Our ($Q=1.3$)} & \textbf{Phase} & \textbf{95.34} & \textbf{94.91} & \textbf{16} \\ \hline
\multicolumn{5}{|c|}{\textbf{ResNet-20}} \\ \hline
\cite{rmp-snn} & Rate & 91.47 & 91.42 & 2048  \\
\cite{TSC} & Temporal & 91.47 & 91.42 & 2048  \\
\cite{optimal_deng} & Rate & 92.14 & 90.89 & 128  \\
\cite{pre_training_iclr}$^\dag$ & Rate & 91.77 & 92.24 & 32  \\
\cite{spikeconverter} & Rate & 91.47 & 91.47 & 16  
\\
\textbf{Ours ($Q=1.3$)} & \textbf{Phase} & \textbf{92.67} & \textbf{92.61} & \textbf{16}\\ \hline
\multicolumn{5}{|c|}{\textbf{ResNet-34}} \\ \hline
 \textbf{Ours ($Q=1.3$)} & \textbf{Phase} & \textbf{95.59} & \textbf{94.84} & \textbf{16} \\ \hline
\end{tabular}%
}\vspace{1mm}
\footnotesize{\dag Conversion-aware ANN training is used}
\end{table}

\begin{table}[!ht]
\caption{Comparisons on Accuracy and Required Timestep on CIFAR-100}
\label{tab:cifar100}
\resizebox{\columnwidth}{!}{%
\begin{tabular}{|lllll|}
\hline
\multicolumn{1}{|l|}{\textbf{Model}} & \multicolumn{1}{l|}{\textbf{Coding}} & \multicolumn{1}{l|}{\textbf{ANN (\%)}} & \multicolumn{1}{l|}{\textbf{SNN (\%)}} & \multicolumn{1}{l|}{\textbf{$T$}}  \\ \hline
\multicolumn{5}{|c|}{\textbf{VGG-16}} \\ \hline
\cite{rmp-snn} & Rate & 71.22 & 70.93 & 2048 \\
\cite{TSC} & Temporal & 71.22 & 70.97 & 2048  \\
\cite{t2fsnn} & Temporal & - & 68.79 & 80  \\
\cite{optimal_deng} & Rate & 71.22 & 70.47 & 128  \\
\cite{pre_training_iclr}$^\dag$ & Rate & 76.28 & 77.02 & 32  \\
\cite{spikeconverter} & Rate & 71.22 & 71.22 & 16  \\
\textbf{Ours ($Q=1.3$)} & \textbf{Phase} & \textbf{72.14} & \textbf{71.85} & \textbf{16}  \\ \hline
\multicolumn{5}{|c|}{\textbf{ResNet-18}} \\ \hline
\cite{pre_training_iclr}$^\dag$ & Rate & 79.62 & 79.62 & 32  \\
\textbf{Ours ($Q=1.3$)} & \textbf{Phase} & \textbf{75.96} & \textbf{76.04} & \textbf{16}  
\\ \hline
\multicolumn{5}{|c|}{\textbf{ResNet-20}} \\ \hline
\cite{rmp-snn} & Rate & 68.72 & 67.82 & 2048  \\
\cite{TSC} & Temporal & 68.72 & 68.18 & 2048  \\
\cite{pre_training_iclr}$^\dag$ & Rate & 69.64 & 69.82 & 32  \\
\cite{spikeconverter} & Rate & 68.72 & 68.69 & 16  \\
\textbf{Ours ($Q=1.3$)} & \textbf{Phase} & \textbf{68.87} & \textbf{68.08} & \textbf{16}\\  \hline
\multicolumn{5}{|c|}{\textbf{ResNet-34}} \\ \hline
\textbf{Ours ($Q=1.3$)} & \textbf{Phase} & \textbf{77.48} & \textbf{76.29} & \textbf{16}  \\ \hline
\end{tabular}%
}\vspace{1mm}
\footnotesize{\dag Conversion-aware ANN training is used}
\end{table}

\begin{table}[!ht]
\caption{Comparisons on Accuracy and Required Timestep on ImageNet}
\label{tab:Imagenet}
\resizebox{\columnwidth}{!}{%
\begin{tabular}{|lllll|}
\hline
\multicolumn{1}{|l|}{\textbf{Model}} & \multicolumn{1}{l|}{\textbf{Coding}} & \multicolumn{1}{l|}{\textbf{ANN (\%)}} & \multicolumn{1}{l|}{\textbf{SNN (\%)}} & \multicolumn{1}{l|}{\textbf{$T$}}  \\ \hline
\multicolumn{5}{|c|}{\textbf{VGG-16}} \\ \hline
\cite{rmp-snn} & Rate & 73.49 & 73.09 & 2560  \\
\cite{TSC} & Temporal & 73.49 & 73.46 & 2560  \\
\cite{optimal_deng} & Rate & 73.47 & 71.06 & 128  \\
\cite{pre_training_iclr}$^\dag$ & Rate & 74.29 & 73.97 & 128  \\
\cite{spikeconverter} & Rate & 73.49 & 73.47 & 16  \\
\textbf{Ours ($Q=1.2$)} & \textbf{Phase} & \textbf{73.36} & \textbf{72.92} & \textbf{24} \\ \hline
\multicolumn{5}{|c|}{\textbf{ResNet-18}} \\ \hline
\textbf{Ours ($Q=1.2$)} & \textbf{Phase} & \textbf{69.75} & \textbf{68.73} & \textbf{24}  \\ \hline
\multicolumn{5}{|c|}{\textbf{ResNet-34}} \\ \hline
\cite{rmp-snn} & Rate & 70.64 & 69.89 & 4096  \\
\cite{TSC} & Temporal & 70.64 & 69.93 & 4096  \\
\cite{pre_training_iclr}$^\dag$ & Rate & 74.32 & 73.15 & 128  \\
\cite{spikeconverter} & Rate & 70.64 & 70.57 & 16  \\
\textbf{Ours ($Q=1.2$)} & \textbf{Phase} & \textbf{73.31} & \textbf{72.21} & \textbf{24}  \\ \hline
\end{tabular}%
}\vspace{1mm}
\footnotesize{\dag Conversion-aware ANN training is used}
\end{table}

RMP-SNN~\cite{rmp-snn} has reduced the conversion loss by using a soft-reset neuron model, and Deep SNN~\cite{TSC} has performed classification through only one spike based on TTFS. 
These two previous studies show high accuracy but require the longest timestep ($\geq 2048$).
The authors in T2FSNN~\cite{t2fsnn}, which uses temporal coding with an exponential decaying kernel, have reduced the timestep to 80 but shows relatively low accuracy.
One can set a different threshold for each layer~\cite{optimal_deng}, so that a smaller accuracy loss can be achieved compared to T2FSNN~\cite{t2fsnn}.
Recently, the authors in~\cite{pre_training_iclr} have successfully reduced the timestep to $32$ while achieving high accuracy by training an ANN with a new customized activation function preventing the conversion loss.
SpikeConverter~\cite{spikeconverter} has shown that the rate coding can be done with $T=16$ without the conversion-aware training.
The authors in~\cite{spikeconverter} have divided the operation of neurons into accumulation and spike generation phases.
Thanks to the techniques described in Section~\ref{sec:reduce_loss}, our one-spike SNN requires only $T=16$ for CIFAR datasets and $T=24$ for ImageNet ({\it fast inference}).
{Even with the minimum timestep (2\textsuperscript{nd} to the best on ImageNet), the one-spike SNN achieves negligible accuracy loss, i.e., 0.35\%, 0.55\% and 0.85\% on average on CIFAR-10, CIFAR-100, and ImageNet, respectively, without placing any constraints to ANN training ({\it accurate and simple conversion}).}

\subsection{Energy Efficiency}\label{sec:energy}
SNNs are energy efficient owing to their event-driven computing characteristic.
Unlike ANNs that require both multiplications and accumulations, SNNs only require additions.
However, if the number of spikes generated during the inference is large, SNNs could be less energy efficient than ANNs.
{

 For the fair comparison, energy costs are estimated based on 32-bit floating-point arithmetic units in 45nm CMOS technology~\cite{horowitz} using the same method employed in the prior work~\cite{diet-snn,onetimestep_allyouneed}.
 } 
One MAC operation consumes 5.11$\times$ more energy, i.e., $4.6$pJ~($E_{MAC}$), than one addition operation, i.e., $0.9$pJ~($E_{addition}$).
Thus, we estimate the relative energy efficiency of SNNs (denoted as $\alpha$) compared to ANNs by using the ratio between the number of MACs in ANN ($OP_{ANN}^l$) and the number of additions in SNN at layer $l$.
\begin{equation}\label{eq:energy}
\alpha = \frac{E_{MAC}\cdot\sum_{l=1}^{N_L} {OP_{ANN}^l}}{E_{addition}\cdot\sum_{l=1}^{N_L}{(SpikeRate_l \cdot OP_{ANN}^l)}},
\end{equation}
where $SpikeRate_l$ is the number of spikes during timestep $T$ of each neuron. 
The number of additions at each SNN layer is approximated by $SpikeRate_l \cdot {OP}_{ANN}^l$.
A higher spike rate translates to higher energy consumption, thus if $SpikeRate_l$ becomes greater than 5.11, the energy consumption will be higher than that of ANN.
Therefore, minimizing the number of spikes as in our single-spike approximation is important to improve the energy efficiency.

Table~\ref{tab:energy} compares the SNN accuracy and energy improvement over ANNs ($\alpha$) between the one-spike SNN and other SNNs with post-training that uses the same energy analysis provided in Eq.~(\ref{eq:energy}).
The prior work~\cite{diet-snn,onetimestep_allyouneed} use the rate coding as an activation encoding scheme and reduce the timestep $T$ by allowing additional post-training epochs.
The proposed one-spike SNN shows comparable energy improvement with 3.92$\sim$5.21\% higher classification accuracy without any post-training.

\subsection{GCN-to-SNN Conversion}

\begin{table}[t]
\caption{Comparisons on Accuracy and Energy Efficiency Between One-spike SNN and Other SNNs with Post-training}
\label{tab:energy}
\resizebox{\columnwidth}{!}{%
\begin{tabular}{|clllll|}
\hline
\multicolumn{1}{|l|}{\textbf{Network}} & \multicolumn{1}{l|}{\textbf{Model}} & \multicolumn{1}{l|}{\textbf{ANN (\%)}} & \multicolumn{1}{l|}{\textbf{SNN (\%)}} & \multicolumn{1}{l|}{\textbf{$T$}} & \textbf{$\alpha$} \\ \hline
\multicolumn{6}{|c|}{\textbf{CIFAR-10}} \\ \hline
\multicolumn{1}{|c|}{{VGG-16}} & \cite{diet-snn} & 93.72 & 92.70 & 5 & 12.4 \\
\multicolumn{1}{|c|}{} & \cite{onetimestep_allyouneed} & 94.10 & 93.05 & 1 & 33 \\
\multicolumn{1}{|c|}{} & \textbf{Ours ($Q=1.3$)} & \textbf{93.9} & \textbf{93.76} & \textbf{16} & \textbf{14.13} \\ \hline
\multicolumn{1}{|l|}{ResNet-18} & \textbf{Ours ($Q=1.3$)} & \textbf{95.34} & \textbf{94.91} & \textbf{16} & \textbf{17.03} \\ \hline
\multicolumn{1}{|c|}{{ResNet-20}} & \cite{diet-snn} & 92.79 & 91.78 & 5 & 6.3 \\
\multicolumn{1}{|c|}{} &  \cite{onetimestep_allyouneed} & 93.34 & 91.10 & 1 & 16.32 \\ 
\multicolumn{1}{|c|}{} & \textbf{Ours ($Q=1.3$)} & \textbf{92.67} & \textbf{92.61} & \textbf{16} & \textbf{13.72} \\ \hline
\multicolumn{1}{|l|}{ResNet-34} & \textbf{Ours ($Q=1.3$) } & \textbf{95.59} & \textbf{94.84} & \textbf{16} & \textbf{17.26} \\ \hline
\multicolumn{6}{|c|}{\textbf{CIFAR-100}} \\ \hline
\multicolumn{1}{|c|}{{VGG-16}} & \cite{diet-snn} & 71.82 & 69.67 & 5 & 12.1 \\
\multicolumn{1}{|c|}{} & \cite{onetimestep_allyouneed} & 72.46 & 70.15 & 1 & 29.24 \\
\multicolumn{1}{|c|}{} & \textbf{Ours ($Q=1.3$)} & \textbf{72.14} & \textbf{71.85} & \textbf{16} & \textbf{13.23} \\ \hline
\multicolumn{1}{|l|}{ResNet-18} & \textbf{Ours ($Q=1.3$)} & \textbf{75.96} & \textbf{76.04} & \textbf{16} & \textbf{15.01} \\ \hline
\multicolumn{1}{|c|}{{ResNet-20}} & \cite{diet-snn} & 64.64 & 64.07 & 5 & 6.6 \\
\multicolumn{1}{|c|}{} & \cite{onetimestep_allyouneed} & 65.90 & 63.30 & 1 & 15.35 \\
\multicolumn{1}{|c|}{} & \textbf{Ours ($Q=1.3$)} & \textbf{68.87} & \textbf{68.08} & \textbf{16} & \textbf{11.56} \\ \hline
\multicolumn{1}{|l|}{ResNet-34} & \textbf{Ours ($Q=1.3$)} & \textbf{77.48} & \textbf{76.29} & \textbf{16} & \textbf{16.9} \\ \hline
\multicolumn{6}{|c|}{\textbf{ImageNet}} \\ \hline
\multicolumn{1}{|c|}{{VGG-16}} & \cite{diet-snn} & 70.08 & 69.00 & 5 & 11.7 \\
\multicolumn{1}{|c|}{} & \cite{onetimestep_allyouneed} & 70.08 & 67.71 & 1 & 24.61 \\
\multicolumn{1}{|c|}{} & \textbf{Ours ($Q=1.2$)} & \textbf{73.36} & \textbf{72.92} & \textbf{24} & \textbf{12.45} \\ \hline
\multicolumn{1}{|l|}{ResNet-18} & \textbf{Ours ($Q=1.2$)} & \textbf{69.75} & \textbf{68.73} & \textbf{24} & \textbf{4.59} \\ \hline
\multicolumn{1}{|l|}{ResNet-34} & \textbf{Ours ($Q=1.2$)} & \textbf{73.31} & \textbf{72.21} & \textbf{24} & \textbf{6.76} \\ \hline
\end{tabular}%
}
\end{table}

\begin{table}[t]
\caption{Accuracy Relative Energy Efficiency of Converting GCNs to One-spike SNNs}
\label{tab:GCN}
\centering
\scalebox{1.1}{
\begin{tabular}{|c|c|c|c|c|c|}
\hline
                  & \textbf{20NG}   & \textbf{R52}    & \textbf{R8}     & \textbf{Ohsumed} & \textbf{MR} \\ \hline
ANN (\%)              & 86.27          & 93.49          & 97.06          & 68.42 &75.93          \\ \hline
\textbf{SNN (\%)} & \textbf{85.90} & \textbf{92.24} & \textbf{96.44} & \textbf{66.95} & \textbf{75.09}  \\ 
\hline
\textbf{\textbf{$\alpha$}} & \textbf{5.91} & \textbf{11.11} & \textbf{6.59} & \textbf{6.24} & \textbf{4.53}  \\ 
\hline
\end{tabular}}
\end{table}

To verify the generality of our ANN-to-SNN conversion method, we selected a graph convolutional network (GCN)~\cite{gcn} as another benchmark.
GCNs are designed on graph structures which consist of a collection of nodes (or vertices), edges, and edge weights. 
The convolution operation in a GCN is a linear combination of the node and neighboring feature vectors.
The authors in~\cite{textgcn} have utilized the GCN to learn representations in text data. 
Each document is represented as a graph, where each node is a word, and edges are connected based on some similarity measure between the words. 
Since a GCN model in~\cite{textgcn} uses a ReLU-based activation function, the pre-trained GCN can be converted to the one-spike SNN without any modifications.

As reported in Table~\ref{tab:GCN}, we converted Text GCNs~\cite{textgcn} to SNNs with $T=16$ and $Q=1.3$ on five benchmarks, i.e., 20-Newsgroups (20NG), R52 and R8 of Reuters 21578, Ohsumed, and Movie Review (MR). 
Text GCN consists of two layers, and the computation of each layer is divided into two steps.
The first step is matrix multiplication with the trained weight matrix ({\it message passing}), and the second step is matrix multiplication with the adjacency matrix ({\it aggregation}).
Each step in GCN is converted as one SNN layer making an one-spike SNN with four layers in total.
{ As a result, direct GCN-to-SNN conversions were successful with an average accuracy loss of 0.9\% while improving the energy efficiency by 4.53$\sim$ 11.11$\times$.}
Since none of the prior work has shown the conversion results on GCNs, we only compared with the ANN baseline.

\section{Discussion}

\subsection{Weight Quantization on One-spike SNNs}

Experimental results discussed in Section~\ref{sec:result} are based on one-spike SNNs with unquantized weights, i.e., FP32 weights, and is not suitable at operating on mobile devices that runs on tight energy budget.
Since each activation value in an original ANN is approximated by single-spike in an one-spike SNN, there could be a concern that the accuracy significantly degrades if weights are quantized. 
As shown in Table~\ref{tab:Quan}, however, our conversion results in negligible accuracy loss even when weights are quantized by 8-bit integers.
By utilizing Eq.~(\ref{eq:energy}), the relative energy efficiency over ANNs when weights are quantized by INT8 ($\alpha_{INT8}$) is computed and reported in the last column of Table~\ref{tab:Quan}. 
{
Since $E_{MAC}$ is $0.2$pJ and $E_{addition}$ is $0.03$pJ for INT8 arithmetic units in 45nm~\cite{horowitz}, the energy efficiency of the quantized SNN model is 1.42$\times$ on average better than that of the SNN model with FP32 weights. } 
Therefore, the proposed ANN-to-SNN conversion process does not hurt accuracy of the original ANN model even with weight quantization to INT8, while achieving much higher energy efficiency suitable at running one-spike SNNs on mobile devices.

\begin{table}[]
\caption{Classification Accuracy and Relative Energy Efficiency ($\alpha_{INT8}$) of One-spike SNNs with Weights Quantized at INT8 
}
\label{tab:Quan}
\resizebox{\columnwidth}{!}{%
\begin{tabular}{|llllll|}
\hline
\multicolumn{1}{|l|}{\textbf{Model}} & \multicolumn{1}{l|}{\textbf{$T$}} & \multicolumn{1}{l|}{\textbf{$Q$}} & \multicolumn{1}{l|}{\textbf{FP32 (\%)}} & \multicolumn{1}{l|}{\textbf{INT8 (\%)}} & $\alpha_{INT8}$ \\ \hline
\multicolumn{6}{|c|}{\textbf{CIFAR-10}} \\ \hline
\multicolumn{1}{|l|}{\textbf{VGG-16}} & 16 & 1.3 & 93.76 & 93.55 & 18.42 \\ \cline{1-1}
\multicolumn{1}{|l|}{\textbf{ResNet-18}} & 16 & 1.3 & 94.91 & 94.93 & 22.22 \\ \cline{1-1}
\multicolumn{1}{|l|}{\textbf{ResNet-20}} & 16 & 1.3 & 92.61 & 92.55 & 20.56 \\ \cline{1-1}
\multicolumn{1}{|l|}{\textbf{ResNet-34}} & 16 & 1.3 & 94.84 & 94.78 & 25.87 \\ \hline
\multicolumn{6}{|c|}{\textbf{CIFAR-100}} \\ \hline
\multicolumn{1}{|l|}{\textbf{VGG-16}} & 16 & 1.3 & 71.85 & 71.97 & 17.26 \\ \cline{1-1}
\multicolumn{1}{|l|}{\textbf{ResNet-18}} & 16 & 1.3 & 76.04 & 75.82 & 19.58 \\ \cline{1-1}
\multicolumn{1}{|l|}{\textbf{ResNet-20}} & 16 & 1.3 & 68.08 & 68.42 & 17.37 \\ \cline{1-1}
\multicolumn{1}{|l|}{\textbf{ResNet-34}} & 16 & 1.3 & 76.29 & 76.19 & 25.29 \\ \hline
\multicolumn{6}{|c|}{\textbf{ImageNet}} \\ \hline
\multicolumn{1}{|l|}{\textbf{VGG-16}} & 24 & 1.2 & 72.92 & 72.85 & 18.55 \\ \cline{1-1}
\multicolumn{1}{|l|}{\textbf{ResNet-18}} & 24 & 1.2 & 68.73 & 68.62 & 6.73 \\ \cline{1-1}
\multicolumn{1}{|l|}{\textbf{ResNet-34}} & 24 & 1.2 & 72.21 & 72.05 & 10.01 \\ \hline
\end{tabular}%
}
\end{table}

\subsection{Single-Spike Approximation on Input Data}\label{sec:input_apprx}

\begin{table}[t!]
\caption{Accuracy of One-spike SNN When Single-spike Approximation is Applied at Input Data}
\label{tab:singleinput}
\resizebox{\columnwidth}{!}{%
\begin{tabular}{|llllll|}
\hline
\multicolumn{1}{|l|}{\textbf{Model}} & \multicolumn{1}{l|}{\textbf{$T$}} & \multicolumn{1}{l|}{\textbf{$Q$}} & \multicolumn{1}{l|}{\textbf{ANN (\%)}} & \multicolumn{1}{l|}{\textbf{SNN (\%)}} & \textbf{$\alpha$} \\ \hline
\multicolumn{6}{|c|}{\textbf{CIFAR-10}} \\ \hline
\multicolumn{1}{|l|}{\textbf{VGG-16}} & 16 & 1.3 & 93.90 & 76.97 & 14.45 \\ \cline{1-1}
\multicolumn{1}{|l|}{\textbf{ResNet-18}} & 16 & 1.3 & 95.34 & 72.99 & 17.67 \\ \cline{1-1}
\multicolumn{1}{|l|}{\textbf{ResNet-20}} & 16 & 1.3 & 92.67 & 56.57 & 14.56 \\ \cline{1-1}
\multicolumn{1}{|l|}{\textbf{ResNet-34}} & 16 & 1.3 & 95.59 & 80.00 & 17.73 \\ \hline
\multicolumn{6}{|c|}{\textbf{CIFAR-100}} \\ \hline
\multicolumn{1}{|l|}{\textbf{VGG-16}} & 16 & 1.3 & 72.14 & 31.87 & 14.11 \\ \cline{1-1}
\multicolumn{1}{|l|}{\textbf{ResNet-18}} & 16 & 1.3 & 75.96 & 36.49 & 16.15 \\ \cline{1-1}
\multicolumn{1}{|l|}{\textbf{ResNet-20}} & 16 & 1.3 & 68.87 & 30.66 & 12.16 \\ \cline{1-1}
\multicolumn{1}{|l|}{\textbf{ResNet-34}} & 16 & 1.3 & 77.48 & 39.91 & 17.27 \\ \hline
\multicolumn{6}{|c|}{\textbf{ImageNet}} \\ \hline
\multicolumn{1}{|l|}{\textbf{VGG-16}} & 24 & 1.2 & 73.36 & 45.74 & 14.10 \\ \cline{1-1}
\multicolumn{1}{|l|}{\textbf{ResNet-18}} & 24 & 1.2 & 69.75 & 16.36 & 10.50 \\ \cline{1-1}
\multicolumn{1}{|l|}{\textbf{ResNet-34}} & 24 & 1.2 & 73.31 & 27.22 & 11.32 \\ \hline
\end{tabular}%
}
\end{table}

The major difference between the temporal coding and the proposed one-spike SNN is that we allow binary coding (multiple spikes) for the input data.
All intermediate layers enjoy the benefit of single-spike approximation which minimizes the energy consumption.
One may think that applying the single-spike approximation to input data will maximize the energy efficiency.
{As shown in Table~\ref{tab:singleinput}, however, the average accuracy loss by applying the single-spike approximation on input data is 22.74\%, 38.88\% and 42.36\% on CIFAR-10, CIFAR-100 and ImageNet, respectively.}
Single-spike approximation on input data is definitely more energy efficient than the binary coding, but the accuracy is greatly sacrificed.
To increase accuracy, precision should be increased by reducing the base $Q$ and increasing timestep $T$ significantly.
As will be described in Section~\ref{sec:noise_spike}, the latency per inference is proportional to the timestep $T$.
Therefore, we stick to the binary coding for encoding the input layer in the one-spike SNN.

\subsection{Limitations}
\subsubsection{Applicable to ReLU-based ANNs}
Our neuron model is tailored for mimicking the ReLU activation function making our ANN-to-SNN conversion limited to ANNs with the ReLU function.
Therefore, all ANN benchmarks in this paper including GCNs are based on ReLU.
If we carefully model a neuron model that approximates a different type of non-linear activation functions, it is expected that ANNs with other types of nonlinear functions can be converted as well, which remains as our future work.

\subsubsection{Latency due to Noise Spikes in Phase Coding}\label{sec:noise_spike}
Phase coding reduces the required timestep $T$ by assigning a weight $Q^{-t}$ to each phase $t$, but cannot immediately propagate spikes to next layers due to noise spikes~\cite{phasecoding}.
Noise spikes prevent the output spike from occurring at the current phase or cause a non-intended spike.
For example, if the membrane potential due to pre-synaptic spikes at phase $t+w$ is greater than $Q^{w}V_{\theta}$, the spike should have generated at phase $t$, not $t+w$ ({\it late spike}).
On the contrary, assume that the output spike is generated at phase $t$.
If the potential update by a negative value happens at a later phase, the output spike at phase $t$ should not happen ({\it false spike}). 
Thus, it is necessary to wait for some phases, i.e., wait timestep $w$, to reduce the computation error due to noise spikes.
Thus, the latency of spike generation at each layer becomes $w$.

We tested multiple wait timesteps $w$ at various $Q$ values when converting to our one-spike SNN.
\begin{table*}[t!]
\caption{Accuracy of One-spike SNN by Varying Wait Timestep $w$ and Base $Q$}
\label{tab:accuracybywait}
\resizebox{\textwidth}{!}{%
\begin{tabular}{|c|cccccccccccccc|c|}
\hline
 &
  \multicolumn{1}{c|}{} &
  \multicolumn{1}{c|}{} &
  \multicolumn{1}{c|}{} &
  \multicolumn{11}{c|}{\textbf{SNN Accuracy ($\%$) by Wait Timestep $w$}} &
   \\ \hline
\textbf{Dataset} &
  \multicolumn{1}{c|}{\textbf{Model}} &
  \multicolumn{1}{c|}{\textbf{T}} &
  \multicolumn{1}{c|}{\textbf{Q}} &
  \multicolumn{1}{c|}{\textbf{4}} &
  \multicolumn{1}{c|}{\textbf{6}} &
  \multicolumn{1}{c|}{\textbf{8}} &
  \multicolumn{1}{c|}{\textbf{10}} &
  \multicolumn{1}{c|}{\textbf{12}} &
  \multicolumn{1}{c|}{\textbf{14}} &
  \multicolumn{1}{c|}{\textbf{16}} &
  \multicolumn{1}{c|}{\textbf{18}} &
  \multicolumn{1}{c|}{\textbf{20}} &
  \multicolumn{1}{c|}{\textbf{22}} &
  \textbf{24} &
  \textbf{ANN} ($\%$) \\ \hline
\textbf{CIFAR-10} &
  \textbf{VGG-16} &
  16 &
  1.3 &
  84.15 &
  91.47 &
  93.56 &
  93.68 &
  93.75 &
  93.83 &
  93.76 &
  - &
  - &
  - &
  - &
  93.90 \\
\textbf{} &
  \textbf{ResNet-18} &
  16 &
  1.3 &
  93.13 &
  94.52 &
  95.00 &
  95.00 &
  94.95 &
  94.98 &
  94.91 &
  - &
  - &
  - &
  - &
  95.34 \\
\textbf{} &
\textbf{ResNet-20} &
  16 &
  1.3 &
  90.47 &
  91.85 &
  92.37 &
  92.37 &
  92.63 &
  92.55 &
  92.61 &
  - &
  - &
  - &
  - &
  92.67 \\
\textbf{} &
  \textbf{ResNet-34} &
  16 &
  1.3 &
  93.63 &
  94.56 &
  94.68 &
  94.83 &
  94.84 &
  94.79 &
  94.84 &
  - &
  - &
  - &
  - &
  95.59 \\ \hline
\textbf{CIFAR-100} &
  \textbf{VGG-16} &
  16 &
  1.3 &
  55.52 &
  62.23 &
  68.20 &
  71.68 &
  72.06 &
  71.78 &
  71.85 &
  - &
  - &
  - &
  - &
  72.14 \\
\textbf{} &
  \textbf{ResNet-18} &
  16 &
  1.3 &
  69.04 &
  74.38 &
  75.30 &
  75.99 &
  75.84 &
  75.93 &
  76.04 &
  - &
  - &
  - &
  - &
  75.96 \\
\textbf{} &
\textbf{ResNet-20} &
  16 &
  1.3 &
  60.52 &
  66.13 &
  67.79 &
  68.01 &
  68.02 &
  68.15 &
  68.08 &
  - &
  - &
  - &
  - &
  68.87 \\
\textbf{} &
  \textbf{ResNet-34} &
  16 &
  1.3 &
  71.84 &
  75.84 &
  76.20 &
  76.38 &
  76.36 &
  76.41 &
  76.29 &
  - &
  - &
  - &
  - &
  77.48 \\ \hline
\textbf{ImageNet} &
  \textbf{VGG-16} &
  24 &
  1.2 &
  22.51 &
  50.87 &
  62.26 &
  66.01 &
  69.16 &
  72.52 &
  73.00 &
  72.91 &
  72.95 &
  72.91 &
  72.92 &
  73.36 \\
 &
  \textbf{ResNet-18} &
  24 &
  1.2 &
  5.00 &
  4.89 &
  11.76 &
  47.32 &
  67.82 &
  68.76 &
  68.65 &
  68.70 &
  68.69 &
  68.72 &
  68.73 &
  69.75 \\
 &
  \textbf{ResNet-34} &
  24 &
  1.2 &
  14.57 &
  18.69 &
  27.79 &
  57.77 &
  71.11 &
  72.07 &
  72.14 &
  72.13 &
  72.10 &
  72.24 &
  72.21 &
  73.31 \\ \hline
\end{tabular}%
}
\end{table*}
Table~\ref{tab:accuracybywait} summarizes the accuracy of one-spike SNN by varying wait timestep $w$ and base $Q$.
As presented, conversion via small $Q$ requires a longer wait timestep $w$ to avoid accuracy loss due to noise spikes.
{For example, the VGG-16 model for the CIFAR datasets has $Q=1.3$, so $w$ needs to be greater than 10. }
However, VGG-16 for ImageNet has $Q=1.2$ where $w$ needs to be greater than 16.
If $w$ is set above a critical value, it only results in subtle differences in accuracy.
According to our experiments, implementing one-spike SNNs with lower $Q$ in each model showed less accuracy loss, but longer $T$ and $w$ were required.
Longer $T$ is required to reduce the underflow error for a small $Q$, and longer $w$ is required to prevent noise spikes to happen in single-spike approximation.

\begin{figure}[t]
    \centering 
    \includegraphics[scale=0.24]{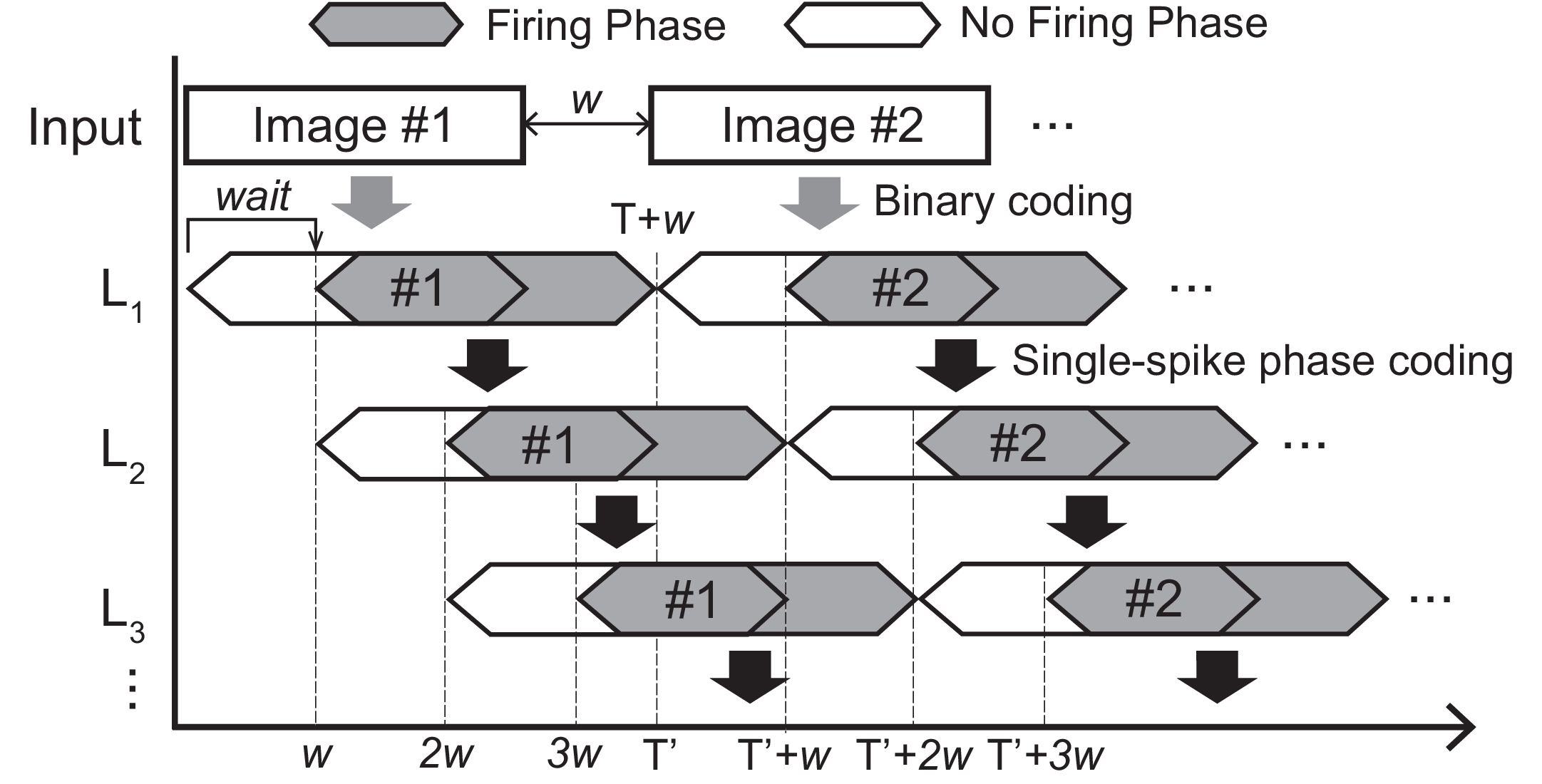}
    \caption{A pipelined execution of SNN layers when multiple images are provided for the inference.}
    \label{fig:pipeline}
\end{figure}

Additional latency is relatively high ($w/T$) when processing a single image.
However, it becomes negligible when the number of images $N_{image}$ to be processed is large enough, as we can pipeline the computation of each layer (Fig.~\ref{fig:pipeline}).
Computation of one SNN layer is done by $T+w$, while the computation of the next SNN layer can start after waiting for $w$ phases.
The next input image is fed into the SNN at every $T+w$.
For $N_{image}$ to be processed, the total latency becomes $N_{image}\cdot(T+w)+ w\cdot(N_L-1)$, where $N_L$ is the total number of layers.
If $N_{image}$ is large, the first term becomes dominant which removes the latency overhead due to wait timestep $w$ (i.e., $w\cdot(N_L-1)$).
We varied $w$ from 8 to 16 with $Q\geq1.2$, and for the ImageNet classification, the inference time per image converged to $(T+w)=40$ when $T=24$.

\subsection{Comparison to Prior Work}
Compared to the rate coding, the phase coding generally presents a smaller conversion loss with a smaller timestep. 
However, as the rate coding requires a larger number of spikes, energy efficiency is worse than one-spike SNN.
To increase the energy efficiency of the rate coding, the timestep must be reduced by post-training~\cite{diet-snn,onetimestep_allyouneed}.
Temporal coding, which generates only one spike, has good energy efficiency.
However, a longer timestep ($>80$) is required to maintain high conversion accuracy~\cite{TSC,t2fsnn}.
The exponential decaying kernel in~\cite{t2fsnn} is similar to our method of changing a $Q$ value.
However, we were able to achieve higher accuracy with smaller timestep thanks to applying binary coding at the input layer and the proper choice of $Q$ and $w$.
The accuracy degradation was 42.36\% on average on ImageNet when the input data is encoded by a single spike as in temporal coding (Section~\ref{sec:input_apprx}). 
Compared to the conventional phase coding, 
our contribution is to increase energy efficiency by minimizing the spike rate.
Similar to ours, the authors in~\cite{logarithimic_coding} have exploited a single-spike approximation, but it fails to minimize conversion error and only shallow ANNs have been tested.
In our experiments, we minimize the conversion loss even with the single-spike approximation and tested on a wide range of benchmarks including GCNs.

\section{Conclusion}
In this paper, we presented an efficient ANN-to-SNN conversion method using a novel single-spike phase coding.
To reduce conversion error, we proposed two techniques: threshold shift for round-off approximation and manipulation of base $Q$.
{Our single-spike approximation minimizes the addition operations during inference and increases the energy efficiency by $4.59\sim17.26\times$ compared to ANNs without post-training the converted SNNs.}
More importantly, our method does not impose any constraints on training ANNs for the conversion process.
This significant energy saving was achieved with negligible accuracy loss tested on a variety of ANN benchmarks: CNNs on CIFAR-10, CIFAR-100 and ImageNet, and GCNs on five datasets.

\bibliographystyle{IEEEtran}
\bibliography{reference}

\begin{thebibliography}{10}
\providecommand{\url}[1]{#1}
\csname url@samestyle\endcsname
\providecommand{\newblock}{\relax}
\providecommand{\bibinfo}[2]{#2}
\providecommand{\BIBentrySTDinterwordspacing}{\spaceskip=0pt\relax}
\providecommand{\BIBentryALTinterwordstretchfactor}{4}
\providecommand{\BIBentryALTinterwordspacing}{\spaceskip=\fontdimen2\font plus
\BIBentryALTinterwordstretchfactor\fontdimen3\font minus \fontdimen4\font\relax}
\providecommand{\BIBforeignlanguage}[2]{{%
\expandafter\ifx\csname l@#1\endcsname\relax
\typeout{** WARNING: IEEEtran.bst: No hyphenation pattern has been}%
\typeout{** loaded for the language `#1'. Using the pattern for}%
\typeout{** the default language instead.}%
\else
\language=\csname l@#1\endcsname
\fi
#2}}
\providecommand{\BIBdecl}{\relax}
\BIBdecl

\bibitem{Alexnet}
A.~Krizhevsky, I.~Sutskever, and G.~E. Hinton, ``Imagenet classification with deep convolutional neural networks,'' in \emph{Proc. of NeurIPS}, 2012.

\bibitem{gpt3}
T.~Brown, B.~Mann, N.~Ryder, M.~Subbiah, J.~D. Kaplan, P.~Dhariwal, A.~Neelakantan, P.~Shyam, G.~Sastry, A.~Askell \emph{et~al.}, ``Language models are few-shot learners,'' in \emph{Proc. of NeurIPS}, vol.~33, 2020, pp. 1877--1901.

\bibitem{video_analytics}
J.~Ngiam, A.~Khosla, M.~Kim, J.~Nam, H.~Lee, and A.~Y. Ng, ``Multimodal deep learning,'' in \emph{Proc. of ICML}, 2011.

\bibitem{evaluate_energy_ann}
D.~Li, X.~Chen, M.~Becchi, and Z.~Zong, ``Evaluating the energy efficiency of deep convolutional neural networks on cpus and gpus,'' in \emph{Proc. of BDCloud}.\hskip 1em plus 0.5em minus 0.4em\relax IEEE, 2016, pp. 477--484.

\bibitem{nature_roy}
K.~Roy, A.~Jaiswal, and P.~Panda, ``Towards spike-based machine intelligence with neuromorphic computing,'' \emph{Nature}, vol. 575, no. 7784, pp. 607--617, 2019.

\bibitem{izh_model}
E.~M. Izhikevich, ``Which model to use for cortical spiking neurons?'' \emph{IEEE Transactions on Neural Networks}, vol.~15, no.~5, pp. 1063--1070, Sept. 2004.

\bibitem{hh_model}
D.~A. McCormick, Y.~Shu, and Y.~Yu, ``{Hodgkin and Huxley} model — still standing?'' \emph{Nature}, vol. 445, no. E1-E2, Jan. 2007.

\bibitem{loihi}
M.~Davies, N.~Srinivasa, T.-H. Lin, G.~Chinya, Y.~Cao, S.~H. Choday, G.~Dimou, P.~Joshi, N.~Imam, S.~Jain, Y.~Liao, C.-K. Lin, A.~Lines, R.~Liu, D.~Mathaikutty, S.~McCoy, A.~Paul, J.~Tse, G.~Venkataramanan, Y.-H. Weng, A.~Wild, Y.~Yang, and H.~Wang, ``Loihi: A neuromorphic manycore processor with on-chip learning,'' \emph{IEEE Micro}, vol.~38, no.~1, pp. 82--99, 2018.

\bibitem{truenorth}
F.~Akopyan, J.~Sawada, A.~Cassidy, R.~Alvarez-Icaza, J.~Arthur, P.~Merolla, N.~Imam, Y.~Nakamura, P.~Datta, G.-J. Nam, B.~Taba, M.~Beakes, B.~Brezzo, J.~B. Kuang, R.~Manohar, W.~P. Risk, B.~Jackson, and D.~S. Modha, ``Truenorth: Design and tool flow of a 65 mw 1 million neuron programmable neurosynaptic chip,'' \emph{IEEE Transactions on Computer-Aided Design of Integrated Circuits and Systems}, vol.~34, no.~10, pp. 1537--1557, 2015.

\bibitem{spinnaker}
S.~B. Furber, D.~R. Lester, L.~A. Plana, J.~D. Garside, E.~Painkras, S.~Temple, and A.~D. Brown, ``Overview of the spinnaker system architecture,'' \emph{IEEE Transactions on Computers}, vol.~62, no.~12, pp. 2454--2467, 2013.

\bibitem{surrogate_train_aaai2021}
H.~Zheng, Y.~Wu, L.~Deng, Y.~Hu, and G.~Li, ``Going deeper with directly-trained larger spiking neural networks,'' in \emph{Proc. of AAAI}, 2021.

\bibitem{surrogate_Train_nips2021}
W.~Fang, Z.~Yu, Y.~Chen, T.~Huang, T.~Masquelier, and Y.~Tian, ``Deep residual learning in spiking neural networks,'' in \emph{Proc. of NeurIPS}, 2021.

\bibitem{Diehl}
P.~U. Diehl and M.~Cook, ``Unsupervised learning of digit recognition using spike-timing-dependent plasticity,'' \emph{Frontiers in Computational Neuroscience}, vol.~9, p.~99, Aug. 2015.

\bibitem{spilinc}
G.~Srinivasan, P.~Panda, and K.~Roy, ``{SpiLinC}: spiking liquid-ensemble computing for unsupervised speech and image recognition,'' \emph{Frontiers in Neuroscience}, vol.~12, p. 524, 2018.

\bibitem{Adaptive-interlink}
S.~Hwang, J.~Lee, and J.~Kung, ``Adaptive input-to-neuron interlink development in training of spike-based liquid state machines,'' in \emph{Proc. of ISCAS}, 2021, pp. 1--5.

\bibitem{rmp-snn}
B.~Han, G.~Srinivasan, and K.~Roy, ``Rmp-snn: Residual membrane potential neuron for enabling deeper high-accuracy and low-latency spiking neural network,'' in \emph{Proc. of CVPR}, 2020, pp. 13\,555--13\,564.

\bibitem{TSC}
B.~Han and K.~Roy, ``Deep spiking neural network: Energy efficiency through time based coding,'' in \emph{Proc. of ECCV}.\hskip 1em plus 0.5em minus 0.4em\relax Springer, 2020, pp. 388--404.

\bibitem{optimal_deng}
S.~Deng and S.~Gu, ``Optimal conversion of conventional artificial neural networks to spiking neural networks,'' in \emph{Proc. of ICLR}, 2020.

\bibitem{spikeconverter}
F.~Liu, W.~Zhao, Y.~Chen, Z.~Wang, and L.~Jiang, ``Spikeconverter: An efficient conversion framework zipping the gap between artificial neural networks and spiking neural networks,'' in \emph{Proc. of AAAI}, 2022.

\bibitem{phasecoding}
J.~Kim, H.~Kim, S.~Huh, J.~Lee, and K.~Choi, ``Deep neural networks with weighted spikes,'' \emph{Neurocomputing}, vol. 311, pp. 373--386, 2018.

\bibitem{temporal-pattern-coding}
B.~Rueckauer and S.-C. Liu, ``Temporal pattern coding in deep spiking neural networks,'' in \emph{Proc. of IJCNN}, 2021.

\bibitem{diet-snn}
N.~Rathi and K.~Roy, ``Diet-snn: A low-latency spiking neural network with direct input encoding and leakage and threshold optimization,'' \emph{IEEE Transactions on Neural Networks and Learning Systems}, 2021.

\bibitem{onetimestep_allyouneed}
S.~S. Chowdhury, N.~Rathi, and K.~Roy, ``Towards ultra low latency spiking neural networks for vision and sequential tasks using temporal pruning,'' in \emph{Proc. of ECCV}, 2022.

\bibitem{pre_training_iclr}
T.~Bu, W.~Fang, J.~Ding, P.~Dai, Z.~Yu, and T.~Huang, ``Optimal ann-snn conversion for high-accuracy and ultra-low-latency spiking neural networks,'' in \emph{Proc. of ICLR}, 2021.

\bibitem{pretraining_ijcai}
J.~Ding, Z.~Yu, Y.~Tian, and T.~Huang, ``Optimal ann-snn conversion for fast and accurate inference in deep spiking neural networks,'' in \emph{Proc. of IJCAI}, 2021.

\bibitem{layer-norm}
P.~Diehl, D.~Neil, J.~Binas, M.~Cook, S.~Liu, and M.~Pfeiffer, ``Fast-classifying, high-accuracy spiking deep networks through weight and threshold balancing,'' in \emph{Proc. of IJCNN}, 2015.

\bibitem{TTFS}
B.~Rueckauer and S.-C. Liu, ``Conversion of analog to spiking neural networks using sparse temporal coding,'' in \emph{Proc. of ISCAS}.\hskip 1em plus 0.5em minus 0.4em\relax Institute of Electrical and Electronics Engineers, 2018, pp. 1--5.

\bibitem{t2fsnn}
S.~Park, S.~Kim, B.~Na, and S.~Yoon, ``T2fsnn: deep spiking neural networks with time-to-first-spike coding,'' in \emph{Proc. of DAC}, 2020, pp. 1--6.

\bibitem{logarithimic_coding}
M.~Zhang, Z.~Gu, N.~Zheng, D.~Ma, and G.~Pan, ``Efficient spiking neural networks with logarithmic temporal coding,'' \emph{IEEE Access}, vol.~8, pp. 98\,156--98\,167, 2020.

\bibitem{channel-wise_norm}
S.~Kim, S.~Park, B.~Na, and S.~Yoon, ``Spiking-yolo: spiking neural network for energy-efficient object detection,'' in \emph{Proc. of AAAI}, 2020.

\bibitem{horowitz}
M.~Horowitz, ``1.1 computing's energy problem (and what we can do about it),'' in \emph{Proc. of ISSCC}, 2014.

\bibitem{t2fsnn_hardware}
D.~Lew, K.~Lee, and J.~Park, ``A time-to-first-spike coding and conversion aware training for energy-efficient deep spiking neural network processor design,'' in \emph{Proc. of DAC}, 2022, pp. 265--270.

\bibitem{gcn}
T.~N. Kipf and M.~Welling, ``Semi-supervised classification with graph convolutional networks,'' in \emph{Proc. of ICLR}, 2017.

\bibitem{textgcn}
L.~Yao, C.~Mao, and Y.~Luo, ``Graph convolutional networks for text classification,'' in \emph{Proc. of AAAI}, 2019.

\end{thebibliography}

\end{document}